\documentclass{neus2025}

\usepackage{hyperref}
\UseRawInputEncoding

\usepackage{url}
\usepackage{booktabs}
\usepackage{lineno}
\usepackage{array}
\usepackage{listings}
\usepackage{amssymb,amsmath}%,amsthm}       % 
\usepackage{amsfonts}      % 
\usepackage{verbatim}
\usepackage{graphicx}
\usepackage{color}
\usepackage{tikz}
\usetikzlibrary{patterns}
\usetikzlibrary{calc}
\usepackage{algorithm}
\usepackage[noend]{algorithmic}
\usepackage{enumitem}

\newcommand{\convexpath}[2]{
  [   
  create hullcoords/.code={
    \global\edef\namelist{#1}
    \foreach [count=\counter] \nodename in \namelist {
      \global\edef\numberofnodes{\counter}
      \coordinate (hullcoord\counter) at (\nodename);
    }
    \coordinate (hullcoord0) at (hullcoord\numberofnodes);
    \pgfmathtruncatemacro\lastnumber{\numberofnodes+1}
    \coordinate (hullcoord\lastnumber) at (hullcoord1);
  },
  create hullcoords
  ]
  ($(hullcoord1)!#2!-90:(hullcoord0)$)
  \foreach [
  evaluate=\currentnode as \previousnode using \currentnode-1,
  evaluate=\currentnode as \nextnode using \currentnode+1
  ] \currentnode in {1,...,\numberofnodes} {
    let \p1 = ($(hullcoord\currentnode) - (hullcoord\previousnode)$),
    \n1 = {atan2(\y1,\x1) + 90},
    \p2 = ($(hullcoord\nextnode) - (hullcoord\currentnode)$),
    \n2 = {atan2(\y2,\x2) + 90},
    \n{delta} = {Mod(\n2-\n1,360) - 360}
    in 
    {arc [start angle=\n1, delta angle=\n{delta}, radius=#2]}
    -- ($(hullcoord\nextnode)!#2!-90:(hullcoord\currentnode)$) 
  }
}

\definecolor{darkblue}{rgb}{0, 0, 0.5}
\hypersetup{colorlinks=true, citecolor=darkblue, linkcolor=darkblue, urlcolor=darkblue}

\title[Benchmarking coherence-driven inference]{Benchmarking graph construction by large language models for coherence-driven inference}
\usepackage{times}
 \author{%
  \Name{Steve Huntsman} \Email{steve.huntsman@cynnovative.com}\\
  \and
  \Name{Jewell Thomas} \Email{jewell.thomas@cynnovative.com}\\
 }

\begin{document}

\maketitle

\begin{abstract}
We devise an algorithm to generate propositions that objectively instantiate graphs supporting coherence-driven inference. We also benchmark the ability of large language models (LLMs) to reconstruct coherence graphs from (a simple transformation of) propositions expressed in natural language, with promising results from a single prompt to reasoning-optimized LLMs. For example, \texttt{o1/3/4-mini} achieve perfect reconstruction half of the time on sparse graphs. Coherence-driven inference on consistency evaluations by LLMs may advance machine cognition capabilities.
\end{abstract}

% \begin{keywords}
%   coherence, large language model, neurosymbolic
% \end{keywords}

\section{\label{sec:introduction}Introduction}

Classical \emph{coherence-driven inference} (CDI) \citep{thagard1998coherence,thagard2002coherence,blokpoel2024theoretical} models many forms of cognition as solving a constraint satisfaction problem. In this approach, propositions and their consistency relations are encoded in a weighted graph $G$ as in Figures \ref{fig:cdi}, \ref{fig:melianDialogue}, and \ref{fig:minuteMystery}. Up to an irrelevant constant that varies among formulations, the \emph{coherence} of $U \subseteq V(G)$ is $-\sum_{u \in U, v \not \in U} A_{uv}$, where $V(G)$ is the vertex set of $G$ and $A$ is the weighted adjacency matrix of $G$. (As is common in the literature, we often assume $A_{uv} \in \{\pm 1\}$ in this paper; our benchmarking approach handles this case naturally.) The problem of maximizing coherence is thus an instance for the matrix $-A$ of the $\mathbf{APX}$-complete MAX-CUT problem \citep{khot2007optimal,moore2011nature,gartner2012approximation,lee2021classifying}. The objective is to cut the most negative weights in $G$ with a bipartition into accepted (= estimated true) and rejected (= estimated false) propositions. 
\footnote{Near-optimal \emph{Rashomon} solutions \citep{semenova2022existence} can become optimal if the graph is perturbed. CDI can account for these by averaging over a Gibbs measure for coherence to indicate probabilities of acceptance.} 
\footnote{In general, context matters for assigning acceptance and rejection to a computed bipartition. The most common approach is to prioritize data. For example, direct observations and firm prior conclusions carry more weight than secondhand reports and tentative prior conclusions. See \citet{blokpoel2024theoretical} for a discussion of this point and remarks about implementing a data priority scheme in the present context. (Implementing a data priority scheme is straightforward in the more general computational context outlined in \S \ref{sec:computationGeneral}: just introduce appropriately weighted [or ``hard''] single-variable clauses into a weighted SAT instance.) A natural criterion \emph{in vacuo} is to prioritize the part of the bipartition whose induced subgraph has greater normalized internal coherence.}

\begin{figure}[htbp]
    \centering   
        \begin{tikzpicture}[scale=1.0]
            % Draw partition first
            \coordinate (A) at (30:1.5);
            \coordinate (B) at (150:1.5);
            \coordinate (C) at (270:1.5);
            \draw[thick,black!0,fill=black,opacity=0.2] \convexpath{A,C}{5mm};
            \draw[thick,black!0,fill=black,opacity=0.2](B) circle (5mm) node {};
            % Labels
            \node (q1) at (90:1.5) {{\color{blue}$+1$}};
            \node (q2) at (210:1.5) {{\color{red}$-1$}};
            \node (q3) at (330:1.5) {{\color{red}$-1$}};
            % Draw nodes
            \node [draw,circle,fill=white,minimum size=6.5mm] (a) at (A) {$a$};
            \node [draw,circle,fill=white,minimum size=6.5mm] (b) at (B) {$b$};
            \node [draw,circle,fill=white,minimum size=6.5mm] (c) at (C) {$c$};
            % Draw edges
            \foreach \from/\to in {
                a/b}
                \draw[ultra thick, blue] (\from) to (\to);
            \foreach \from/\to in {
                a/c,b/c}
                \draw[ultra thick, red, dashed] (\from) to (\to);
            % Draw cut
            \coordinate (cut1) at (70:1.5);
            \coordinate (cut2) at (230:1.5);
            \draw[line width=5pt, color=gray] (cut1) to (cut2);
            % % Annotation
            % \node[align=center] (coh) at (270:2) {Coherence = \\ -(({\color{blue}+1}) + ({\color{red}-1})) = 0};
        \end{tikzpicture}
        \quad
        \begin{tikzpicture}[scale=1.0]
            % Draw partition first
            \coordinate (A) at (30:1.5);
            \coordinate (B) at (150:1.5);
            \coordinate (C) at (270:1.5);
            \draw[thick,black!0,fill=black,opacity=0.2] \convexpath{A,B}{5mm};
            \draw[thick,black!0,fill=black,opacity=0.2](C) circle (5mm) node {};
            % Labels
            \node (q1) at (90:1.5) {{\color{blue}$+1$}};
            \node (q2) at (210:1.5) {{\color{red}$-1$}};
            \node (q3) at (330:1.5) {{\color{red}$-1$}};
            % Draw nodes
            \node [draw,circle,fill=white,minimum size=6.5mm] (a) at (A) {$a$};
            \node [draw,circle,fill=white,minimum size=6.5mm] (b) at (B) {$b$};
            \node [draw,circle,fill=white,minimum size=6.5mm] (c) at (C) {$c$};
            % Draw edges
            \foreach \from/\to in {
                a/b}
                \draw[ultra thick, blue] (\from) to (\to);
            \foreach \from/\to in {
                a/c,b/c}
                \draw[ultra thick, red, dashed] (\from) to (\to);
            % Draw cut
            \coordinate (cut1) at (-10:1.5);
            \coordinate (cut2) at (190:1.5);
            \draw[line width=5pt, color=gray] (cut1) to (cut2);
            % % Annotation
            % \node[align=center] (coh) at (270:2) {Coherence = \\ -(({\color{red}-1}) + ({\color{red}-1})) = 2};
        \end{tikzpicture}
    \caption{
    Left: the coherence graph $G$ associated with the propositions $a: X \text{ is hot}$, $b: X \text{ is bright}$, and $c: X \text{ is cold and dark}$. {\color{blue}Consistent (solid blue)} and {\color{red}inconsistent (red dashed)} pairs of vertices (= propositions) respectively get {\color{blue}positive} and {\color{red}negative} weights. Also shown is the cut $\{\{a, c\},\{b\}\}$ whose coherence is $-(A_{ab} + A_{bc}) = -(({\color{blue}+1}) + ({\color{red}-1})) = 0$, where $A$ is the weighted adjacency matrix of $G$. Right: the associated coherence graph $G$ with the cut $\{\{a, b\},\{c\}\}$ whose coherence is $-(A_{ac} + A_{bc}) = -(({\color{red}-1}) + ({\color{red}-1})) = 2$. CDI amounts to accepting and rejecting data according to a cut $\{U, V(G) - U\}$ that maximizes the coherence objective $-\sum_{u \in U, v \not \in U} A_{uv}$. The cut on the right is optimal, separating propositions according to their inconsistencies.
    }
    \label{fig:cdi}
\end{figure}

\begin{figure}[htbp]
  \centering
    \resizebox{.68\textwidth}{!}{%
        % true 2
        \begin{tikzpicture}[
          xscale=4, % Adjust scaling as needed
          yscale=4,
        ]
        \node[align=left,font=\footnotesize] (a) at (0,0) {
        \# Athenians \\
        - $p1$: The strong dominate and the weak endure. \\
        - $p2$: Internal rebellion is a greater threat to Athens than external rivals like Lacedaemon. \\
        - $p3$: Subjugating Melos would strengthen Athenian security and expand the Athenian empire. \\
        - $p4$: Melian neutrality would show Athenian weakness. \\
        - $p5$: Hope is risky and unreliable, especially for those with few resources like the Melians. \\
        - $p6$: Both gods and men support ruling wherever possible. \\
        - $p7$: The Lacedaemonians prioritize security and are unlikely to intervene in support of Melos. \\
        
        \\
        
        \# Melians \\
        - $p8$: Justice and fairness protect weaker states like Melos. \\
        - $p9$: Athenian aggression risks creating more enemies for Athens and alienating neutral states. \\
        - $p10$: Resisting Athens is about honor and survival for the Melians. \\
        - $p11$: War's unpredictability offers hope to the Melians against stronger foes like Athens. \\
        - $p12$: The gods and the Lacedaemonians will support Melos due to justice and kinship with the Melians.
             };
        \end{tikzpicture}
    }
    \resizebox{.28\textwidth}{!}{%
        % first time: Melian12props.txt_20250513_091304.pkl
        \begin{tikzpicture}
          % Define styles
          \tikzset{
            node/.style={circle, draw=black, fill=white, minimum size=1cm},
          }
        
          % Nodes
          \node[node] (nodep1) at (2.8978, .77645) {$p1$};
          \node[node] (nodep10) at (.77645, -2.8978) {$p10$};
          \node[node] (nodep11) at (2.1213, -2.1213) {$p11$};
          \node[node] (nodep12) at (2.8978, -.77645) {$p12$};
          \node[node] (nodep2) at (2.1213, 2.1213) {$p2$};
          \node[node] (nodep3) at (.77645, 2.8978) {$p3$};
          \node[node] (nodep4) at (-.77645, 2.8978) {$p4$};
          \node[node] (nodep5) at (-2.1213, 2.1213) {$p5$};
          \node[node] (nodep6) at (-2.8978, .77645) {$p6$};
          \node[node] (nodep7) at (-2.8978, -.77645) {$p7$};
          \node[node] (nodep8) at (-2.1213, -2.1213) {$p8$};
          \node[node] (nodep9) at (-.77645, -2.8978) {$p9$};
        
          % Edges
          \draw[solid, color={rgb,1:red,0.500;green,0.000;blue,0.500}, opacity=0.00, line width=3pt] (nodep1) -- (nodep10);
          \draw[solid, color={rgb,1:red,0.500;green,0.000;blue,0.500}, opacity=0.00, line width=3pt] (nodep1) -- (nodep11);
          \draw[solid, color={rgb,1:red,0.500;green,0.000;blue,0.500}, opacity=0.00, line width=3pt] (nodep1) -- (nodep12);
          \draw[solid, color={rgb,1:red,0.500;green,0.000;blue,0.500}, opacity=0.00, line width=3pt] (nodep1) -- (nodep2);
          \draw[solid, color={rgb,1:red,0.000;green,0.000;blue,1.000}, opacity=1.00, line width=3pt] (nodep1) -- (nodep3);
          \draw[solid, color={rgb,1:red,0.500;green,0.000;blue,0.500}, opacity=0.00, line width=3pt] (nodep1) -- (nodep4);
          \draw[solid, color={rgb,1:red,0.500;green,0.000;blue,0.500}, opacity=0.00, line width=3pt] (nodep1) -- (nodep5);
          \draw[solid, color={rgb,1:red,0.200;green,0.000;blue,0.800}, opacity=0.60, line width=3pt] (nodep1) -- (nodep6);
          \draw[solid, color={rgb,1:red,0.500;green,0.000;blue,0.500}, opacity=0.00, line width=3pt] (nodep1) -- (nodep7);
          \draw[solid, color={rgb,1:red,0.500;green,0.000;blue,0.500}, opacity=0.00, line width=3pt] (nodep1) -- (nodep8);
          \draw[solid, color={rgb,1:red,0.500;green,0.000;blue,0.500}, opacity=0.00, line width=3pt] (nodep1) -- (nodep9);
          \draw[solid, color={rgb,1:red,0.200;green,0.000;blue,0.800}, opacity=0.60, line width=3pt] (nodep10) -- (nodep11);
          \draw[solid, color={rgb,1:red,0.500;green,0.000;blue,0.500}, opacity=0.00, line width=3pt] (nodep10) -- (nodep12);
          \draw[solid, color={rgb,1:red,0.500;green,0.000;blue,0.500}, opacity=0.00, line width=3pt] (nodep10) -- (nodep2);
          \draw[solid, color={rgb,1:red,0.500;green,0.000;blue,0.500}, opacity=0.00, line width=3pt] (nodep10) -- (nodep3);
          \draw[solid, color={rgb,1:red,0.500;green,0.000;blue,0.500}, opacity=0.00, line width=3pt] (nodep10) -- (nodep4);
          \draw[solid, color={rgb,1:red,0.200;green,0.000;blue,0.800}, opacity=0.60, line width=3pt] (nodep10) -- (nodep5);
          \draw[solid, color={rgb,1:red,0.500;green,0.000;blue,0.500}, opacity=0.00, line width=3pt] (nodep10) -- (nodep6);
          \draw[solid, color={rgb,1:red,0.500;green,0.000;blue,0.500}, opacity=0.00, line width=3pt] (nodep10) -- (nodep7);
          \draw[solid, color={rgb,1:red,0.250;green,0.000;blue,0.750}, opacity=0.50, line width=3pt] (nodep10) -- (nodep8);
          \draw[solid, color={rgb,1:red,0.500;green,0.000;blue,0.500}, opacity=0.00, line width=3pt] (nodep10) -- (nodep9);
          \draw[solid, color={rgb,1:red,0.500;green,0.000;blue,0.500}, opacity=0.00, line width=3pt] (nodep11) -- (nodep12);
          \draw[solid, color={rgb,1:red,0.500;green,0.000;blue,0.500}, opacity=0.00, line width=3pt] (nodep11) -- (nodep2);
          \draw[solid, color={rgb,1:red,0.500;green,0.000;blue,0.500}, opacity=0.00, line width=3pt] (nodep11) -- (nodep3);
          \draw[solid, color={rgb,1:red,0.500;green,0.000;blue,0.500}, opacity=0.00, line width=3pt] (nodep11) -- (nodep4);
          \draw[dashed, color={rgb,1:red,0.700;green,0.000;blue,0.300}, opacity=0.40, line width=3pt] (nodep11) -- (nodep5);
          \draw[solid, color={rgb,1:red,0.500;green,0.000;blue,0.500}, opacity=0.00, line width=3pt] (nodep11) -- (nodep6);
          \draw[solid, color={rgb,1:red,0.500;green,0.000;blue,0.500}, opacity=0.00, line width=3pt] (nodep11) -- (nodep7);
          \draw[solid, color={rgb,1:red,0.500;green,0.000;blue,0.500}, opacity=0.00, line width=3pt] (nodep11) -- (nodep8);
          \draw[solid, color={rgb,1:red,0.500;green,0.000;blue,0.500}, opacity=0.00, line width=3pt] (nodep11) -- (nodep9);
          \draw[solid, color={rgb,1:red,0.500;green,0.000;blue,0.500}, opacity=0.00, line width=3pt] (nodep12) -- (nodep2);
          \draw[solid, color={rgb,1:red,0.500;green,0.000;blue,0.500}, opacity=0.00, line width=3pt] (nodep12) -- (nodep3);
          \draw[solid, color={rgb,1:red,0.500;green,0.000;blue,0.500}, opacity=0.00, line width=3pt] (nodep12) -- (nodep4);
          \draw[solid, color={rgb,1:red,0.500;green,0.000;blue,0.500}, opacity=0.00, line width=3pt] (nodep12) -- (nodep5);
          \draw[dashed, color={rgb,1:red,0.700;green,0.000;blue,0.300}, opacity=0.40, line width=3pt] (nodep12) -- (nodep6);
          \draw[dashed, color={rgb,1:red,1.000;green,0.000;blue,0.000}, opacity=1.00, line width=3pt] (nodep12) -- (nodep7);
          \draw[solid, color={rgb,1:red,0.000;green,0.000;blue,1.000}, opacity=1.00, line width=3pt] (nodep12) -- (nodep8);
          \draw[solid, color={rgb,1:red,0.500;green,0.000;blue,0.500}, opacity=0.00, line width=3pt] (nodep12) -- (nodep9);
          \draw[solid, color={rgb,1:red,0.350;green,0.000;blue,0.650}, opacity=0.30, line width=3pt] (nodep2) -- (nodep3);
          \draw[solid, color={rgb,1:red,0.500;green,0.000;blue,0.500}, opacity=0.00, line width=3pt] (nodep2) -- (nodep4);
          \draw[solid, color={rgb,1:red,0.500;green,0.000;blue,0.500}, opacity=0.00, line width=3pt] (nodep2) -- (nodep5);
          \draw[solid, color={rgb,1:red,0.500;green,0.000;blue,0.500}, opacity=0.00, line width=3pt] (nodep2) -- (nodep6);
          \draw[solid, color={rgb,1:red,0.000;green,0.000;blue,1.000}, opacity=1.00, line width=3pt] (nodep2) -- (nodep7);
          \draw[solid, color={rgb,1:red,0.500;green,0.000;blue,0.500}, opacity=0.00, line width=3pt] (nodep2) -- (nodep8);
          \draw[solid, color={rgb,1:red,0.200;green,0.000;blue,0.800}, opacity=0.60, line width=3pt] (nodep2) -- (nodep9);
          \draw[solid, color={rgb,1:red,0.000;green,0.000;blue,1.000}, opacity=1.00, line width=3pt] (nodep3) -- (nodep4);
          \draw[solid, color={rgb,1:red,0.500;green,0.000;blue,0.500}, opacity=0.00, line width=3pt] (nodep3) -- (nodep5);
          \draw[solid, color={rgb,1:red,0.250;green,0.000;blue,0.750}, opacity=0.50, line width=3pt] (nodep3) -- (nodep6);
          \draw[solid, color={rgb,1:red,0.500;green,0.000;blue,0.500}, opacity=0.00, line width=3pt] (nodep3) -- (nodep7);
          \draw[solid, color={rgb,1:red,0.500;green,0.000;blue,0.500}, opacity=0.00, line width=3pt] (nodep3) -- (nodep8);
          \draw[solid, color={rgb,1:red,0.300;green,0.000;blue,0.700}, opacity=0.40, line width=3pt] (nodep3) -- (nodep9);
          \draw[solid, color={rgb,1:red,0.500;green,0.000;blue,0.500}, opacity=0.00, line width=3pt] (nodep4) -- (nodep5);
          \draw[solid, color={rgb,1:red,0.500;green,0.000;blue,0.500}, opacity=0.00, line width=3pt] (nodep4) -- (nodep6);
          \draw[solid, color={rgb,1:red,0.450;green,0.000;blue,0.550}, opacity=0.10, line width=3pt] (nodep4) -- (nodep7);
          \draw[solid, color={rgb,1:red,0.500;green,0.000;blue,0.500}, opacity=0.00, line width=3pt] (nodep4) -- (nodep8);
          \draw[solid, color={rgb,1:red,0.300;green,0.000;blue,0.700}, opacity=0.40, line width=3pt] (nodep4) -- (nodep9);
          \draw[solid, color={rgb,1:red,0.500;green,0.000;blue,0.500}, opacity=0.00, line width=3pt] (nodep5) -- (nodep6);
          \draw[solid, color={rgb,1:red,0.500;green,0.000;blue,0.500}, opacity=0.00, line width=3pt] (nodep5) -- (nodep7);
          \draw[solid, color={rgb,1:red,0.500;green,0.000;blue,0.500}, opacity=0.00, line width=3pt] (nodep5) -- (nodep8);
          \draw[solid, color={rgb,1:red,0.500;green,0.000;blue,0.500}, opacity=0.00, line width=3pt] (nodep5) -- (nodep9);
          \draw[solid, color={rgb,1:red,0.500;green,0.000;blue,0.500}, opacity=0.00, line width=3pt] (nodep6) -- (nodep7);
          \draw[solid, color={rgb,1:red,0.500;green,0.000;blue,0.500}, opacity=0.00, line width=3pt] (nodep6) -- (nodep8);
          \draw[solid, color={rgb,1:red,0.500;green,0.000;blue,0.500}, opacity=0.00, line width=3pt] (nodep6) -- (nodep9);
          \draw[solid, color={rgb,1:red,0.500;green,0.000;blue,0.500}, opacity=0.00, line width=3pt] (nodep7) -- (nodep8);
          \draw[solid, color={rgb,1:red,0.500;green,0.000;blue,0.500}, opacity=0.00, line width=3pt] (nodep7) -- (nodep9);
          \draw[solid, color={rgb,1:red,0.500;green,0.000;blue,0.500}, opacity=0.00, line width=3pt] (nodep8) -- (nodep9);
        \end{tikzpicture}
    }
  \caption{Left: The 12 most important propositions in the Melian Dialogue from \citet{thucydides1874history} as extracted and processed by \texttt{GPT-4o}. (NB. Lacedaemon is an ancient name for Sparta.) Right: The median of 30 coherence graphs produced by single prompts to \texttt{o1-mini} along the lines of \S \ref{sec:prompt_practical}. Convergence of the median coherence graph is illustrated in \S \ref{sec:l1_distances}. {\color{blue}Consistent (solid blue)} and {\color{red}inconsistent (red dashed)} pairs of vertices (= propositions) respectively get {\color{blue}positive} and {\color{red}negative} weights, with intermediate colors and transparency indicating the strength of (in)consistency. The optimal cut has parts $\{p1, p2, p3, p4, p6, p7, p9\}$ and $\{p5, p8, p10, p11, p12\}$. Note that $p5$ and $p9$ ``cross the line'' between Athens and Melos in opposite directions: their significance is highlighted in history. A hopeful Melos was brutally conquered by Athens in 416 BCE, during a truce with Sparta. Sparta later defeated Athens in the Peloponnesian War in 405 BCE, removed Athenians from Melos, and resettled surviving Melians there.
  }
  \label{fig:melianDialogue}
\end{figure}

\begin{figure}[htbp]
  \centering
    \resizebox{.28\textwidth}{!}{%
        % true 2

    }
  \caption{Left: Propositions adapted from the first mystery in \citet{ripley1932minute}, along with its solution and six relevant hypotheses. Center: The median of 30 coherence graphs produced by single prompts to \texttt{o1-mini}. Right: The result of omitting solution data from prompts. In each median graph an optimal cut {\color{blue}accepts the key observation $p3$} and {\color{red}rejects hypotheses $p20$, $p23$, and $p25$} that are inconsistent with the solution explaining Butler's guilt. 
  }
  \label{fig:minuteMystery}
\end{figure}

CDI is deeply informed by cognitive science, and experimentally validated by psychological case studies, including assessments of legal reasoning \citep{holyoak1999bidirectional,simon2004third}. It is a computationally credible model for causal inference \citep{thagard2004causal} suited for making decisions about ill-structured problems \citep{frigotto2015explanatory}. Other illustrations of its capacity for high-level reasoning include reaching deliberative consensus \citep{joseph2009coherence}, solving normative inconsistencies \citep{criado2016coherence}, and many complementary examples cited in \citet{blokpoel2024theoretical}. CDI can also explicitly incorporate ethical considerations and provide explanations of its reasoning \citep{yilmaz2017computational,sivaraj2019cogent}. However, mechanisms for automatically generating coherence graphs from natural language have not been developed: coherence graphs have almost always been constructed manually. (A notable exception is the specialized construction of \citet{joseph2009coherence} with deduction from underlying logic.) 

In this paper, we show how to generate a set of propositions expressible in natural language that objectively instantiate a given signed coherence graph. We then use this result to benchmark the ability of large language models (LLMs) \citep{brown2020language} to (re)construct signed coherence graphs, with promising results from a single prompt to models optimized for reasoning: see Figures \ref{fig:example1}-\ref{fig:cohere_graphs} and \S \ref{sec:uncertainty}. Given a ``correct'' coherence graph, the CDI process amounts to running a MAX-CUT solver or analogue thereof, along with conceptually straightforward ancillary tasks. Viewed in this light, our work demonstrates the initial feasibility for a neurosymbolic architecture \citep{sarker2022neuro,marra2024statistical} that combines neural reasoning by an LLM to compile a coherence graph with symbolic reasoning via CDI to reason over the graph. This architecture naturally separates concerns and is unique in the breadth of straightforward applications.

In particular, CDI is suited for ``slow'' or ``system 2'' reasoning with hard computations performed on impoverished representations \citep{kahneman2011thinking}. In contrast, previous neurosymbolic efforts that use LLMs to help with graph coloring \citep{khandelwal2024neurosymbolic} or transform natural language data into a logical formula to be passed to a solver  \citep{olausson2023linc,pan2023logic,ye2024satlm} reason at a low level of abstraction, cannot resolve ambiguity after the representational process, and can suffer from parsing errors \citep{feng2024language}. Meanwhile, benchmarks appropriate for evaluating the performance of such efforts using propositional or first-order logic \citep{zhong2021ar,han2022folio,saparov2023language} are not suitable for evaluating coherence.  

Meanwhile, transformer-based architectures \citep{vaswani2017attention} like LLMs are suited for ``fast'' or ``system 1'' reasoning with easy computations on rich representations \citep{lenat2023getting,bengio2024machine}. Transformers are computationally weak without chain-of-thought (CoT) reasoning, but become more powerful with it \citep{merrill2024expressive}. In principle, using CoT linear in input size allows transformers to simulate automata and recognize regular languages, or emulate a probabilistic Turing machine in quasi-quadratic time \citep{perez2021attention, nowak2024representational}. Transformers using polynomial-size CoT can also solve the problems in $\mathbf{P}$. However, the computation of a transformer with $L$ layers on a prompt of length $n$ can be performed using $O(L \log n)$ bits of memory under common assumptions. As a corollary, multilayer transformers cannot scalably solve the easy problems 2-SAT or Horn-SAT unless $\mathbf{L} = \mathbf{NL}$ \citep{peng2024limitations}. On the other hand, the representational power of multi-head transformer architectures is bounded \emph{below} by $n$-gram models \citep{rajaraman2024transformers}, which perform well in practice \citep{liu2024infini}.

Besides its clean separation of neural and symbolic reasoning, a key advantage and enabler of our approach is that an LLM takes independent and identically distributed samples from some distribution over coherence graphs. Our results show that even individual coherence graphs can be reconstructed with excellent precision and recall. There is a simple explanation for this. Asking an LLM to attempt to logically \emph{interpolate} between pairs of propositions and characterize this process with a numerical rating is more tightly constrained, and therefore less prone to hallucinations and their ilk, than \emph{extrapolating} responses from open-ended prompts. 
% (If a sample reveals clear multimodality, then resampling from each mode and performing inference accordingly may be appropriate.) 
Taking a median of the individual realizations almost inevitably yields robust coherence graphs. This procedure amounts to optimal consensus in the sense of graph edit distance and an optimal rank one approximation in the $L^1$ norm of the matrix whose columns are vectorized weighted adjacency matrices. Furthermore, resampling and taking medians yields a quantitative measure of robustness. 

Observations on realistic examples such as Figure \ref{fig:melianDialogue}---for which convergence is shown in \S \ref{sec:l1_distances}---and Figure \ref{fig:minuteMystery} (for which convergence is not shown, but similar) indicate that the median $L^1$/graph edit distance between medians of $n$ out of $N = 30$ responses to prompts generally decreases rapidly to near zero as a function of $n$. (The decrease is slower in more complex examples such as poker hands. Sharp, late decreases appear to indicate topological ``phase transitions'' in global connectivity that require more capable models or other scaling fixes.) Moreover, while several median coherence graphs generated for (e.g.) Figures \ref{fig:melianDialogue} and \ref{fig:minuteMystery} exhibited some residual variation, our experiments suggest that these generally share an optimal cut (not detailed here, but see the right panel of Figure \ref{fig:minuteMystery}). It is therefore plausible that individual coherence graph realizations are sparse approximations that approximately preserve all of the cuts of a denser, implicit ``Platonic'' (hyper)graph \citep{benczur1996approximating,spielman2011spectral,soma2019spectral}. If this is indeed the case, it could explain connectivity transitions as due to hyperedges being sparsified in uniformly different ways, so that resulting edges vanish when taking a median over a large enough sample.

Our approach holds promise for autonomy that is explainable (by analyzing [near-] optimal cuts); ethical (by including ethical guidelines as data \citep{yilmaz2017computational,sivaraj2019cogent}); reproducible and stable (see, e.g., Figure \ref{fig:minuteMystery} and \S \ref{sec:l1_distances}); versatile (by using any sufficiently capable multimodal or domain-specific model); intrinsically capable of handling abstraction and ambiguity; grounded in a flexible cognitive model; and generalizable by construction (see \S \ref{sec:computationGeneral}). 

The examples we provide are already at scales commensurate with hand-crafted coherence graphs in the literature. As such, the scalability of our approach for many if not most practical purposes is governed by the ability to organize and compose smaller inferences into larger inferences. This amounts to promoting (only) relevant accepted propositions from one instance of inference to the next, whether hierarchically or sequentially. Scaling to ``simply'' larger coherence graphs to achieve more human (or superhuman) cognitive abilities is of questionable utility in light of a fixed-parameter tractable cognition hypothesis articulated in \cite{van2008tractable,van2019cognition}.

The paper is organized as follows. \S \ref{sec:modelingCoherenceGraphs} explains how to produce propositions that naturally correspond to a given coherence graph. \S \ref{sec:experimentalSetup} outlines our experimental setup and \S \ref{sec:results} details the most important results of our experiments. We conclude in \S \ref{sec:conclusion}. Appendices respectively give substantial information on generalizing and computing CDI (\S \ref{sec:computation}); a case study in which \texttt{ChatGPT 4} outperformed a skilled human at assigning consistency ratings to proposition pairs (\S \ref{sec:vincennes}); ANOVA (\S \ref{sec:anova}); modeling and reconstructing under uncertainty (\S \ref{sec:uncertainty}); the effects of using different graph-theoretical algorithms in our benchmarking approach (\S \ref{sec:construction}); a preliminary mechanistic interpretation of how attention mechanisms may underlie our results (\S \ref{sec:cross_encoders}); characteristics of the graphs we generated for benchmarking (\S \ref{sec:graph_characteristics}); an example of how median consensus coherence graphs converge (\S \ref{sec:l1_distances}); our prompt templates (\S \ref{sec:prompt_example} and \S \ref{sec:prompt_practical}), and finally an alternative fidelity characterization (\S \ref{sec:l1_norm_graph_anova}). 

{\bf We plan to release code and data in the coming weeks.}

\section{\label{sec:modelingCoherenceGraphs}Modeling coherence graphs}

Here we describe how to benchmark classical CDI using signed graphs, which approximate more elaborate constructions such as weighted simplicial complexes that may be appropriate for modeling and computing coherence in full generality (for which see \citet{huntsman2024prospects} and \S \ref{sec:computationGeneral}). 

{\definition Recall that a \emph{signed graph} $G_\sigma = (V,E,\sigma)$ is an undirected graph $G = (V,E)$ augmented with a \emph{sign map} $\sigma: E \rightarrow \{\pm 1\}$. A \emph{signed coherence graph} is a signed graph whose vertices are propositions. 
\footnote{
In practice we can use a more expressive framework than (natural language generation applied to) propositional logic, but we use the term \emph{proposition} throughout; Algorithm \ref{alg:ModelCoherenceGraph} involves \emph{bona fide} propositional formulas.
}
Two propositions $v$ and $w$ are \emph{dependent} in $G_\sigma$ if $(v,w) \in E$ (in particular, $v$ and $w$ are assumed to be distinct). Two dependent propositions are \emph{consistent for $\sigma$} (resp., \emph{inconsistent for $\sigma$}) if $\sigma(v,w) = +1$ (resp., $-1$). The left panel of Figure \ref{fig:variables} shows an example.}

\begin{figure}[htbp]
    \centering
        \begin{tikzpicture}[scale=1.0]
            % Draw nodes
            \node [draw,circle,fill=white,minimum size=6.5mm] (a) at (90:1.5) {$a$};
            \node [draw,circle,fill=white,minimum size=6.5mm] (b) at (162:1.5) {$b$};
            \node [draw,circle,fill=white,minimum size=6.5mm] (c) at (234:1.5) {$c$};
            \node [draw,circle,fill=white,minimum size=6.5mm] (d) at (306:1.5) {$d$};
            \node [draw,circle,fill=white,minimum size=6.5mm] (e) at (18:1.5) {$e$};
            % Draw edges
            \foreach \from/\to in {
                a/e,b/c,b/e,c/d}
                \draw[ultra thick, blue] (\from) to (\to);
            \foreach \from/\to in {
                a/b,b/d,d/e}
                \draw[ultra thick, red, dashed] (\from) to (\to);
        \end{tikzpicture}
        \quad        
        \begin{tikzpicture}[scale=1.0]
            % Draw cliques first
            \coordinate (A) at (90:1.5);
            \coordinate (B) at (162:1.5);
            \coordinate (C) at (234:1.5);
            \coordinate (D) at (306:1.5);
            \coordinate (E) at (18:1.5);
            \draw[thick,black!0,fill=black,opacity=0.2] \convexpath{A,E,B}{3mm};
            \draw[thick,black!0,fill=black,opacity=0.2] \convexpath{B,D,C}{3mm};
            \draw[thick,black!0,fill=black,opacity=0.2] \convexpath{E,D}{3mm};
            % Label cliques
            \node (q1) at (90:.75) {$q_1$};
            \node (q2) at (234:.75) {$q_2$};
            \node (q3) at (342:.75) {$q_3$};
            % Draw nodes
            \node [draw,circle,fill=white,minimum size=6.5mm] (a) at (A) {$a$};
            \node [draw,circle,fill=white,minimum size=6.5mm] (b) at (B) {$b$};
            \node [draw,circle,fill=white,minimum size=6.5mm] (c) at (C) {$c$};
            \node [draw,circle,fill=white,minimum size=6.5mm] (d) at (D) {$d$};
            \node [draw,circle,fill=white,minimum size=6.5mm] (e) at (E) {$e$};
            % Draw edges
            \foreach \from/\to in {
                a/e,b/c,b/e,c/d}
                \draw[ultra thick, blue] (\from) to (\to);
            \foreach \from/\to in {
                a/b,b/d,d/e}
                \draw[ultra thick, red, dashed] (\from) to (\to);
        \end{tikzpicture}
        \quad
        \begin{tikzpicture}
            \node[align=left] (a) at (0,0) {$\text{Variables}(a') = \{q_1\}$ \\
            $\text{Variables}(b') = \{q_1,q_2\}$ \\
            $\text{Variables}(c') = \{q_2\}$ \\
            $\text{Variables}(d') = \{q_2,q_3\}$ \\
            $\text{Variables}(e') = \{q_1,q_3\}$ \\
            };
        \end{tikzpicture}
    \caption{
    Left: a signed coherence graph for propositions $\{a,\dots,e\}$, with {\color{blue}consistent pairs $(a,e)$, $(b,c)$, $(b,e)$, and $(c,d)$ indicated by solid blue edges, corresponding to edge weights of $+1$}; and {\color{red}inconsistent pairs $(a,b)$, $(b,d)$, and $(d,e)$ indicated by dashed red edges, corresponding to edge weights of $-1$}. Center: {\color{gray}cliques $q_1$, $q_2$, and $q_3$ (indicated by gray vertex clusters)} yield a minimal clique edge partition (hence cover). Right: each vertex is assigned variables corresponding to the cliques that cover it.
    }
    \label{fig:variables}
\end{figure}

A \emph{consistency oracle} $\varepsilon$ is a function that sends any pair of distinct propositions expressed in natural language to $\{-1,0,1\}$. When $\varepsilon(v,w) = -1$, $0$, and $1$, $v$ and $w$ are respectively \emph{(dependent and) inconsistent for $\varepsilon$}, \emph{independent for $\varepsilon$}, and \emph{(dependent and) consistent for $\varepsilon$}. While LLMs can effectively detect inconsistencies between two propositions (see \citet{huntsman2024prospects,kumar2024nlp}, and \S \ref{sec:vincennes}), we will initially restrict consideration  to synthetic propositions for which the output of any ``reasonable'' consistency oracle (i.e., one in which propositions that express logical formulas in natural language are evaluated in the natural way suggested by the underlying formulas) is practically obvious and unambiguous.

{\definition Given a consistency oracle $\varepsilon$, a signed coherence graph $G_\sigma = (V,E,\sigma)$, a set of propositions $V'$ expressed in natural language, and a bijection $\iota : V' \rightarrow V$, we say that $V'$ \emph{models $G_\sigma$ relative to $\varepsilon$} and write $V' \models_\varepsilon G_\sigma$ if for all distinct $v', w' \in V'$ we have $\varepsilon(v',w') = \sigma_0(\iota(v'),\iota(w'))$, where $\sigma_0$ extends $\sigma$ by the value $0$ on pairs of independent distinct propositions.}

We will give a general method for constructing $V' \models_\varepsilon G_\sigma$ under any reasonable consistency oracle $\varepsilon$, which we informally write $V' \models G_\sigma$. 

For example, taking $G_\sigma$ to be the signed coherence graph in the left panel of Figure \ref{fig:variables}, 
\begin{multline*}
V' = \{a': (q_1 \text{ is } p_1), \quad b': (q_1 \text{ is NOT } p_1 \text{ AND } q_2 \text{ is NOT } p_1), \quad c': (q_2 \text{ is } p_2), \\
d': (q_2 \text{ is } p_1 \text{ AND } q_3 \text{ is } p_1), \quad e': (q_1 \text{ is } p_2 \text{ AND } q_3 \text{ is NOT } p_1)\}
\end{multline*}
where the \emph{(clique) variables} $q_j$ are placeholders for nouns and the \emph{properties} $p_k$ are placeholders for gradable (hence negatable) adjectives (e.g., ``hot,'' ``bright,'' ``loud,'' etc.) and $\iota : a' \mapsto a, \dots, e' \mapsto e$, we can see that $V' \models G_\sigma$. 
A structurally equivalent model $V'' \models G_\sigma$ is
\begin{multline*}
  V'' = \{ a'': (\text{the living room is hot}), \quad
  b'': (\text{the living room is cold and the kitchen is cold}), \\
  c'': (\text{the kitchen is bright}), \quad
  d'': (\text{the kitchen is hot and the bedroom is hot}), \\
  e'': (\text{the living room is bright and the bedroom is cold}) \}.
\end{multline*}
% \begin{align*}
%   V'' = \{ a'': & \ (\text{the living room is hot}), \\
%   b'': & \ (\text{the living room is cold and the kitchen is cold}), \\
%   c'': & \ (\text{the kitchen is bright}), \\
%   d'': & \ (\text{the kitchen is hot and the bedroom is hot}), \\
%   e'': & \ (\text{the living room is bright and the bedroom is cold}) \}.
% \end{align*}
While there are structurally inequivalent sets of propositions that model $G_\sigma$, a set of the form above is extremal in that its clique edge cover and star forest decomposition are minimal: see \S \ref{sec:basicModel}.

\subsection{\label{sec:basicModel}An algorithm for modeling signed coherence graphs}

Here we detail how to construct a set of propositions that model a signed coherence graph.

{\definition A \emph{clique edge cover} of a graph $G$ \citep{erdos1966representation,gross2018graph,conte2020large} is a set of cliques in $G$ whose edges cover $E(G)$. A \emph{star forest decomposition} of $G$ \citep{akiyama1985factors,kottarathil2024graph} is a partition of $E(G)$ into star forests, i.e., forests whose connected components each have only a single vertex of degree greater than 1: see Figure \ref{fig:properties}.}

\begin{figure}[htbp]
    \centering
        \begin{tikzpicture}[scale=1.0]
            % Draw star forests first
            \coordinate (W) at (-1,1);
            \coordinate (X) at (-1,-1);
            \coordinate (Y) at (1,-1);
            \coordinate (Z) at (1,1);
            \draw[black!0,pattern color=white,pattern=north east lines] \convexpath{W,Z}{4mm};
            \draw[black!0,pattern color=white,pattern=north west lines] \convexpath{X,Y}{4mm};
            \draw[black!0,pattern color=white,pattern=horizontal lines] \convexpath{X,Z}{4mm};
            % Draw nodes
            \node [draw,circle,fill=white,minimum size=6.5mm] (w) at (-1,1) {$w$};
            \node [draw,circle,fill=white,minimum size=6.5mm] (x) at (-1,-1) {$x$};
            \node [draw,circle,fill=white,minimum size=6.5mm] (y) at (1,-1) {$y$};
            \node [draw,circle,fill=white,minimum size=6.5mm] (z) at (1,1) {$z$};
            % Draw edges
            \foreach \from/\to in {
                w/x,w/y,y/z}
                \draw[ultra thick, blue] (\from) to (\to);
            \foreach \from/\to in {
                w/z,x/y,x/z}
                \draw[ultra thick, red, dashed] (\from) to (\to);
        \end{tikzpicture}
        \quad
        \begin{tikzpicture}[scale=1.0]
            % Draw star forests first
            \coordinate (W) at (-1,1);
            \coordinate (X) at (-1,-1);
            \coordinate (Y) at (1,-1);
            \coordinate (Z) at (1,1);
            \draw[black!0,pattern color=lightgray,pattern=north east lines] \convexpath{W,Z}{4mm};
            \draw[black!0,pattern color=lightgray,pattern=north west lines] \convexpath{X,Y}{4mm};
            \draw[black!0,pattern color=lightgray,pattern=horizontal lines] \convexpath{X,Z}{4mm};
            % Draw nodes
            \node [draw,circle,fill=white,minimum size=6.5mm] (w) at (-1,1) {$w$};
            \node [draw,circle,fill=white,minimum size=6.5mm] (x) at (-1,-1) {$x$};
            \node [draw,circle,fill=white,minimum size=6.5mm] (y) at (1,-1) {$y$};
            \node [draw,circle,fill=white,minimum size=6.5mm] (z) at (1,1) {$z$};
            % Draw edges
            \foreach \from/\to in {
                w/x,w/y,y/z}
                \draw[ultra thick, blue] (\from) to (\to);
            \foreach \from/\to in {
                w/z,x/y,x/z}
                \draw[ultra thick, red, dashed] (\from) to (\to);
            % Annotate star forests
            \node (A1) at (-.55,1.2) {$1$};
            \node (A0) at (.55,1.2) {$!1$};
            \node (B1) at (-.55,-1.2) {$2$};
            \node (B0) at (.55,-1.2) {$!2$};
            \node (C1) at (-.8,-.5) {$3$};
            \node (C0) at (.8,.5) {$!3$};
        \end{tikzpicture}
        \quad
        \begin{tikzpicture}[scale=1.0]
            % Draw star forests first
            \coordinate (W) at (-1,1);
            \coordinate (X) at (-1,-1);
            \coordinate (Y) at (1,-1);
            \coordinate (Z) at (1,1);
            \draw[black!0,pattern color=lightgray,pattern=north east lines] \convexpath{W,Z}{4mm};
            \draw[black!0,pattern color=lightgray,pattern=north west lines] \convexpath{X,Y}{4mm};
            \draw[black!0,pattern color=lightgray,pattern=north east lines] \convexpath{X,Z}{4mm};
            % Draw nodes
            \node [draw,circle,fill=white,minimum size=6.5mm] (w) at (-1,1) {$w$};
            \node [draw,circle,fill=white,minimum size=6.5mm] (x) at (-1,-1) {$x$};
            \node [draw,circle,fill=white,minimum size=6.5mm] (y) at (1,-1) {$y$};
            \node [draw,circle,fill=white,minimum size=6.5mm] (z) at (1,1) {$z$};
            % Draw edges
            \foreach \from/\to in {
                w/x,w/y,y/z}
                \draw[ultra thick, blue] (\from) to (\to);
            \foreach \from/\to in {
                w/z,x/y,x/z}
                \draw[ultra thick, red, dashed] (\from) to (\to);
            % Annotate star forests
            \node (A1) at (-.55,1.2) {$1$};
            \node (A0) at (.55,1.2) {$!1$};
            \node (B1) at (-.55,-1.2) {$2$};
            \node (B0) at (.55,-1.2) {$!2$};
            \node (C1) at (-.8,-.5) {$1$};
            \node (C0) at (.8,.5) {$!1$};
        \end{tikzpicture}
        \quad
        \begin{tikzpicture}[scale=1.0]
            % Draw star forests first
            \coordinate (W) at (-1,1);
            \coordinate (X) at (-1,-1);
            \coordinate (Y) at (1,-1);
            \coordinate (Z) at (1,1);
            \draw[black!0,pattern color=lightgray,pattern=north east lines] \convexpath{W,Z}{4mm};
            \draw[black!0,pattern color=lightgray,pattern=north west lines] \convexpath{X,Y}{4mm};
            \draw[black!0,pattern color=lightgray,pattern=north west lines] \convexpath{X,Z}{4mm};
            % Draw nodes
            \node [draw,circle,fill=white,minimum size=6.5mm] (w) at (-1,1) {$w$};
            \node [draw,circle,fill=white,minimum size=6.5mm] (x) at (-1,-1) {$x$};
            \node [draw,circle,fill=white,minimum size=6.5mm] (y) at (1,-1) {$y$};
            \node [draw,circle,fill=white,minimum size=6.5mm] (z) at (1,1) {$z$};
            % Draw edges
            \foreach \from/\to in {
                w/x,w/y,y/z}
                \draw[ultra thick, blue] (\from) to (\to);
            \foreach \from/\to in {
                w/z,x/y,x/z}
                \draw[ultra thick, red, dashed] (\from) to (\to);
            % Annotate star forests
            \node (A1) at (-.55,1.2) {$1$};
            \node (A0) at (.55,1.2) {$!1$};
            \node (B1) at (-.55,-1.2) {$2$};
            \node (B0) at (.55,-1.2) {$!2$};
            \node (C1) at (-.8,-.5) {$2$};
            \node (C0) at (.8,.5) {$!2$};
        \end{tikzpicture}
    \caption{
    Left: a clique in a notional signed coherence graph. Center left: the degenerate star forest decomposition of negative edges yields the greatest number of corresponding properties. The notations $n$ and $!n$ respectively indicate inconsistent propositional clauses of the form $(q \Rightarrow p_n)$ and $(q \Rightarrow \lnot p_n)$. For example, if $\Rightarrow p_1$, $\Rightarrow p_2$, and $\Rightarrow p_3$ respectively correspond to ``is hot,'', ``is bright,'' and ``is loud,'' then a natural language expression of modeling propositions is $\{w' : (q \text{ is hot}), \ x' : (q \text{ is bright AND loud}), \ y' : (q \text{ is dark}), \ z' : (q \text{ is cold AND quiet})\}$. Right panels: the two star forest decompositions of minimal size yield the minimal number of corresponding properties. The panel on the right corresponds to $\{w' : (q \text{ is hot}), \ x' : (q \text{ is bright}), \ y' : (q \text{ is dark}), \ z' : (q \text{ is cold AND dark})\}$. 
    }
    \label{fig:properties}
\end{figure}

Note that the sizes of clique edge covers and star forest decompositions are saturated below by the \emph{intersection number} and \emph{star arboricity}, respectively.

{\theorem Given a signed coherence graph $G_\sigma = (V,E,\sigma)$, Algorithm \ref{alg:ModelCoherenceGraph} produces $V' \models G_\sigma$.}

\begin{proof}
It suffices to show that $\varepsilon(v',w') = \sigma_0(\iota(v'),\iota(w'))$ via a case analysis. 

First, suppose that $\sigma_0(\iota(v'),\iota(w')) = 1$: here, we need to show that $v'$ and $w'$ are consistent. The respective formulas $\phi(\iota(v'))$ and $\phi(\iota(w'))$ only share common variables in clauses of the form $(q_j \Rightarrow p(v'))$ and $(q_j \Rightarrow p(w'))$. These formulas and their natural language expressions are both consistent unless $p(v') = \lnot p(w')$, which can only occur if $\sigma_0(\iota(v'),\iota(w')) = -1$, a contradiction.

Next, suppose that $\sigma_0(\iota(v'),\iota(w')) = 0$: here, we need to show that $v'$ and $w'$ are independent. Now $\iota(v')$ and $\iota(w')$ are not in a common clique, so the respective formulas $\phi(\iota(v'))$ and $\phi(\iota(w'))$ do not share any common variables. The required result follows.

Finally, suppose that $\sigma_0(\iota(v'),\iota(w')) = -1$: here, we need to show that $v'$ and $w'$ are inconsistent. Now the formulas $\phi(\iota(v'))$ and $\phi(\iota(w'))$ respectively contain clauses of the form $(q_j \Rightarrow p_k)$ and $(q_j \Rightarrow \lnot p_k)$ for some clique $C_j$ and star forest $F_k$ in the graph induced by $E^-_j$. Thus both the formulas and their natural language expressions are inconsistent. 
\end{proof}

\begin{algorithm}
  \caption{\textsc{ModelCoherenceGraph}$(G_\sigma)$}
  \label{alg:ModelCoherenceGraph}
\begin{algorithmic}[1]
  \REQUIRE Signed coherence graph $G_\sigma$ 
  \STATE Produce a clique edge cover $C = \{C_1,\dots,C_m\}$ of $G_\sigma$
  \FOR{$v \in V = V(G_\sigma)$}
    \STATE $\phi(v) \leftarrow \varnothing$ \hfill \emph{// initialize with empty propositional formula}
  \ENDFOR
  \FOR{$C_j \in C$}
    \STATE $E^-_j \leftarrow \{(v,w) \in E(C_j) : \sigma(v,w) = -1\}$
    \STATE Produce a star forest decomposition $F = \{F_1,\dots,F_n\}$ of the graph induced by $E^-_j$
    \FOR{$F_k \in F$}
      \FOR{each star $S \subseteq F_k$}
        \STATE $r \leftarrow \text{root of } S$
        \STATE $\phi(r) \leftarrow \phi(r) \land (q_j \Rightarrow p_k)$ \hfill \emph{// ``\dots AND $q_j$ has property $p_k$''}
        \FOR{each leaf vertex $\ell \in S$}
          \STATE $\phi(\ell) \leftarrow \phi(\ell) \land (q_j \Rightarrow \lnot p_k)$ \hfill \emph{// ``\dots AND $q_j$ has property NOT($p_k$)''}
        \ENDFOR
      \ENDFOR
    \ENDFOR
  \ENDFOR
  \FOR{$v \in V$}
    \IF{$\phi(v) = \varnothing$}
      \STATE $\phi(v) \leftarrow (q^*_v \Rightarrow p^*_v)$ \hfill \emph{// to avoid triviality; $q^*_v$ only appears in $\phi(v)$}
    \ENDIF
    \STATE $v' \leftarrow \text{natural language expression of } \phi(v)$ \hfill \emph{// see comments above}
    \STATE $\iota(v') \leftarrow v$
  \ENDFOR
  \ENSURE $\{\iota^{-1}(v) : v \in V\} = V' \models G_\sigma$
\end{algorithmic}
\end{algorithm}

Note that Algorithm \ref{alg:ModelCoherenceGraph} does not require a specific construction for clique edge covers or for star forest decompositions. Different constructions affect the numbers of variables and properties: see \S \ref{sec:construction} for different clique edge cover approaches. While the approach of \citet{conte2020large} gives practical linear-time performance for clique edge covers on graphs of thousands to millions of vertices, we deal with smaller graphs and we can therefore employ more pedestrian techniques:
\begin{itemize}[noitemsep]
    \item \texttt{degenerate}: take the degenerate clique edge cover formed by individual edges; 
    \item \texttt{percolation}: find all maximal cliques, then use a constraint solver to optimize a clique edge cover (see \url{https://stackoverflow.com/a/49145938});
    \item \texttt{partition}: initialize $G' = G$, then greedily partition $E(G')$ by iteratively removing edges from the largest clique, as discussed in but avoided by \citet{conte2020large}.
\end{itemize}
Note that the first and last of these techniques actually yield a clique edge partition.

Practical algorithms are in shorter supply for star forest decompositions. Finding small decompositions is $\mathbf{NP}$-complete \citep{hakimi1996star}. An integer linear program for star decompositions (to be aggregated into star forest decompositions) is described in \citet{hajebi2024parameterized}, while \citet{cicalese2021star} gives a linear time (1/2)-approximation algorithm for the largest minimum star size. An arboricity-realizing algorithm such as \citet{gabow1988forests} yields a (1/2)-approximate star arboricity-realizing algorithm by the observation of \citet{alon1992star} that any tree can be covered by two star forests. Here we only use the second of the following techniques:
\begin{itemize}[noitemsep]
    \item take the degenerate star forest decomposition formed by individual edges;
    \item use the (1/2)-approximate algorithm for the largest minimum star size;
    \item repeatedly generate uniformly random edge partitions \citep{stam1983generation} and test that they contain no triangles or simple paths of (edge) length 3 \citep{aider2019star}: this is a viable brute force strategy for connected components on up to about 10 vertices since the corresponding number of partitions---i.e., the Bell number $B_{10}$---is 115975.
\end{itemize}

\subsection{\label{sec:benchmark}Benchmarking coherence}

Because computing coherence amounts to solving MAX-CUT (or MAX-SAT more generally as in \S \ref{sec:computationGeneral}), benchmarking CDI is mainly a question of fidelity of coherence graph reconstruction. Runtime and approximation performance of combinatorial optimization algorithms in \S \ref{sec:computation} may become important, but at scales considered here, exact solutions are always computationally feasible. 
\footnote{
Inferring a coherence graph from a set of propositions is similar to classical natural language inference or textual entailment \citep{korman2018defining}, which involves evaluating the logical relationship between two sentences (see \S \ref{sec:cross_encoders}). However, explanatory coherence describes more general properties of individual propositions and sets of propositions in the context of theory formation. Further, coherence encompasses a range of relations that include deductive entailment but also explanatory, analogical, perceptual, conceptual, and deliberative relations \citep{thagard2002coherence}. 
}

The propositions we generate for benchmarking use a grammar that can represent multiple types of relations and uncertainty (see \S \ref{sec:fuzzy_uncertainty}). We note that the specific syntax used to represent propositions in the benchmark dataset is a matter of convenience, and the benchmark can be adapted to represent synthetic propositions using any syntax that is suitable for a given application. 

We use a ``micro'' $F_1$ score to evaluate fidelity of coherence graph reconstruction for each attempt. This accounts for class imbalance among the values in $\{-1,0,1\}$ for each entry in each synthetic adjacency matrix included in our evaluation task \citep{opitz2024closer}. % In particular, we calculate F1 micro by calculating micro precision as $\frac{\sum \text{TP}}{\sum (\text{TP} + \text{FP})}$, micro recall as $\frac{\sum \text{TP}}{\sum (\text{TP} + \text{FN})}$ where TP, FP, TN, and FP are calculated across all edge types simultaneously. We then summarize F1 micro as $\frac{2 \times \text{Micro Precision} \times \text{Micro Recall}}{\text{Micro Precision} + \text{Micro Recall}}$.
\footnote{
Another approach to gauging performance is to use a distance measure, as in \S \ref{sec:l1_norm_graph_anova}.}

\section{\label{sec:experimentalSetup}Experimental setup}

We include models intentionally designed for reasoning in our evaluation.  We refer to \texttt{o1/3/4-mini} \citep{jaech2024openai,openai2025openai}, \texttt{QwQ-32B} \citep{qwen_team_qwq_2024}, and \texttt{Sky-T1-32B} \citep{sky_t1_2025} as \emph{large reasoning models} (LRMs), following the phrasing of \citet{valmeekam2024llms} to describe models designed to generate long chains of thought at inference time. 
% Table \ref{tab:models}.  
These models have high latency and inference costs \citep{abdin2024phi}. We call \texttt{phi-4} \citep{abdin2024phi} a \emph{small language model} (SLM). We call the remaining models LLMs: \texttt{gemini-2.0-flash-exp} \citep{google_deepmind_gemini_2024}, \texttt{claude-3.5-sonnet} \citep{anthropic_claude_2024}, \texttt{gpt-4o} \citep{hurst2024gpt}, \texttt{llama-3.3-70B} \citep{meta_llama_3_3_2024}, \texttt{llama-4-scout} \citep{meta_llama_4_2025}, and \texttt{gemini-1.5-pro} \citep{pichai_hassabis_2024}. 
% The output of all models was post-processed to extract structured data for evaluation.
For all models, we post-processed outputs to extract structured data for evaluation. 

{\bf NB. We are still benchmarking a DeepSeek model as of this writing. We further plan to evaluate other models (e.g., GPT-5, more recent versions of Gemini and Sonnet and other Llama versions/derivatives), and to report results in an update.}

\label{sec:evalPropositions}{
\subsection{Benchmark generation}
}
%\subsubsection{Coherence graph generation}

To instantiate our benchmark, we sample signed coherence graphs from an Erd\H{o}s-R\'enyi (ER) distribution. We ensure that each sample graph is fully connected by joining sampled graphs to a minimum spanning tree generated for a complete graph of equivalent size with random edge weights. 

For comparison with practical examples, the degree variance goodness-of-fit test in \citet{ouadah2020degree} and \citet{brune2025goodness} fails to reject the ER hypothesis for the (unweighted version of the) graph in Figure \ref{fig:melianDialogue}. However, the same test does reject the ER hypothesis for graphs in Figure \ref{fig:minuteMystery} with $p$-values around $0.001$. Other realistic examples not detailed here also fall into both categories. 

We generate 76 graphs with $5 \le |V| \le 23$, targeting two regimes of edge density $:= |E|/$\scalebox{.75}{$\binom{|V|}{2}$}: \texttt{sparse} has a median edge density of 0.202 and \texttt{dense} has a median edge density of 0.734.

\begin{figure}[t]
  \centering
    \resizebox{.24\textwidth}{!}{%
        % true 1
        \begin{tikzpicture}[
          node distance=2cm, % Adjust as needed
          every node/.style={circle,draw,minimum size=0.7cm,inner sep=0,outer sep=0},
          blue edge/.style={blue, ultra thick},
          red edge/.style={red, ultra thick, dashed},
          xscale=4, % Adjust scaling as needed
          yscale=4,
        ]
        
        % Node positions
        \node (a) at (1.0,4.257474623792601e-09) {$a$};
        \node (h) at (-0.5000000585402762,0.8660253918837754) {$h$};
        \node (b) at (0.9555727839785156,0.29475519465920813) {$b$};
        \node (n) at (-0.7330518352608938,-0.6801727366730055) {$n$};
        \node (l) at (-0.9888308034135964,-0.14904225926382042) {$l$};
        \node (u) at (0.9555727839785156,-0.29475527555122594) {$u$};
        \node (j) at (-0.9009688483100784,0.4338838199533992) {$j$};
        \node (c) at (0.8262387515347354,0.563320104165219) {$c$};
        \node (d) at (0.6234897973824897,0.7818315066171321) {$d$};
        \node (q) at (0.07473030006975624,-0.9972037623345826) {$q$};
        \node (s) at (0.6234896185685554,-0.7818316173114723) {$s$};
        \node (e) at (0.3653409783575216,0.9308737552372659) {$e$};
        \node (g) at (-0.22252094658888455,0.9749279057873659) {$g$};
        \node (m) at (-0.9009688483100784,-0.4338837518338052) {$m$};
        \node (f) at (0.07473012125582204,0.9972037708495318) {$f$};
        \node (o) at (-0.49999990952866424,-0.8660254429734708) {$o$};
        \node (r) at (0.36534100815984394,-0.9308737467223167) {$r$};
        \node (i) at (-0.7330518948655386,0.68017268558331) {$i$};
        \node (k) at (-0.9888308034135964,0.14904232738341439) {$k$};
        \node (p) at (-0.22252100619352927,-0.9749278972724167) {$p$};
        \node (t) at (0.82623881113938,-0.5633199764409804) {$t$};
        
        % Edges
        \draw[blue edge] (a) -- (h);
        \draw[blue edge] (a) -- (b);
        \draw[red edge] (a) -- (n);
        \draw[blue edge] (a) -- (m);
        \draw[blue edge] (h) -- (g);
        \draw[red edge] (h) -- (p);
        \draw[red edge] (b) -- (l);
        \draw[red edge] (b) -- (u);
        \draw[blue edge] (b) -- (j);
        \draw[red edge] (b) -- (t);
        \draw[blue edge] (n) -- (r);
        \draw[blue edge] (n) -- (f);
        \draw[blue edge] (n) -- (p);
        \draw[blue edge] (l) -- (c);
        \draw[red edge] (l) -- (m);
        \draw[blue edge] (l) -- (d);
        \draw[blue edge] (j) -- (f);
        \draw[red edge] (d) -- (q);
        \draw[blue edge] (d) -- (s);
        \draw[red edge] (s) -- (e);
        \draw[red edge] (s) -- (t);
        \draw[blue edge] (s) -- (f);
        \draw[blue edge] (e) -- (g);
        \draw[red edge] (e) -- (m);
        \draw[blue edge] (e) -- (i);
        \draw[red edge] (g) -- (i);
        \draw[red edge] (g) -- (k);
        \draw[blue edge] (f) -- (o);
        \draw[blue edge] (f) -- (r);
        \draw[blue edge] (r) -- (t);
        \draw[blue edge] (i) -- (t);
        
        \end{tikzpicture}
    }
    \resizebox{.24\textwidth}{!}{%
        % reconstruction 1
        \begin{tikzpicture}[
          node distance=2cm, % Adjust as needed
          every node/.style={circle,draw,minimum size=0.7cm,inner sep=0,outer sep=0},
          blue edge/.style={blue, ultra thick},
          red edge/.style={red, ultra thick, dashed},
          xscale=4, % Adjust scaling as needed
          yscale=4,
        ]
        
        % Node positions
        \node (e) at (0.3653409783575216,0.9308737552372659) {$e$};
        \node (g) at (-0.22252094658888455,0.9749279057873659) {$g$};
        \node (i) at (-0.7330518948655386,0.68017268558331) {$i$};
        \node (f) at (0.07473012125582204,0.9972037708495318) {$f$};
        \node (n) at (-0.7330518352608938,-0.6801727366730055) {$n$};
        \node (r) at (0.36534100815984394,-0.9308737467223167) {$r$};
        \node (a) at (1.0,4.257474623792601e-09) {$a$};
        \node (b) at (0.9555727839785156,0.29475519465920813) {$b$};
        \node (h) at (-0.5000000585402762,0.8660253918837754) {$h$};
        \node (m) at (-0.9009688483100784,-0.4338837518338052) {$m$};
        \node (j) at (-0.9009688483100784,0.4338838199533992) {$j$};
        \node (l) at (-0.9888308034135964,-0.14904225926382042) {$l$};
        \node (t) at (0.82623881113938,-0.5633199764409804) {$t$};
        \node (u) at (0.9555727839785156,-0.29475527555122594) {$u$};
        \node (c) at (0.8262387515347354,0.563320104165219) {$c$};
        \node (d) at (0.6234897973824897,0.7818315066171321) {$d$};
        \node (q) at (0.07473030006975624,-0.9972037623345826) {$q$};
        \node (s) at (0.6234896185685554,-0.7818316173114723) {$s$};
        \node (o) at (-0.49999990952866424,-0.8660254429734708) {$o$};
        \node (k) at (-0.9888308034135964,0.14904232738341439) {$k$};
        \node (p) at (-0.22252100619352927,-0.9749278972724167) {$p$};
        
        % Edges
        \draw[blue edge] (e) -- (g);
        \draw[blue edge] (e) -- (i);
        \draw[red edge] (e) -- (m);
        \draw[red edge] (e) -- (s);
        \draw[red edge] (g) -- (i);
        \draw[blue edge] (g) -- (h);
        \draw[red edge] (g) -- (k);
        \draw[blue edge] (i) -- (t);
        \draw[blue edge] (f) -- (n);
        \draw[blue edge] (f) -- (r);
        \draw[blue edge] (f) -- (j);
        \draw[blue edge] (f) -- (o);
        \draw[blue edge] (f) -- (s);
        \draw[blue edge] (n) -- (r);
        \draw[red edge] (n) -- (a);
        \draw[blue edge] (n) -- (p);
        \draw[blue edge] (r) -- (t);
        \draw[blue edge] (a) -- (b);
        \draw[blue edge] (a) -- (h);
        \draw[blue edge] (a) -- (m);
        \draw[blue edge] (b) -- (j);
        \draw[red edge] (b) -- (l);
        \draw[red edge] (b) -- (t);
        \draw[red edge] (b) -- (u);
        \draw[red edge] (h) -- (p);
        \draw[red edge] (m) -- (l);
        \draw[blue edge] (l) -- (c);
        \draw[blue edge] (l) -- (d);
        \draw[red edge] (t) -- (s);
        \draw[red edge] (d) -- (q);
        \draw[blue edge] (d) -- (s);
        
        \end{tikzpicture}
    }
    \resizebox{.49\textwidth}{!}{%
        % true 2
        \begin{tikzpicture}[
          xscale=4, % Adjust scaling as needed
          yscale=4,
        ]
            \node[align=left] (a) at (0,0) {
                $a'$: \texttt{q3 is 0.655*Q AND q4 is 0.7*Q AND q5 is 0.585*Q AND q6 is 0.642*P} \\
                $b'$: \texttt{q3 is 0.698*Q AND q7 is 0.66*Q AND q8 is 0.582*P AND q9 is 0.688*P AND q10 is 0.614*P} \\
                $c'$: \texttt{q11 is 0.57*Q} \\
                $d'$: \texttt{q12 is 0.668*Q AND q13 is 0.672*P AND q14 is 0.656*Q} \\
                $e'$: \texttt{q1 is 0.642*Q AND q15 is 0.346*P AND q16 is 0.619*P} \\
                $f'$: \texttt{q2 is 0.639*Q AND q17 is 0.68*Q AND q18 is 0.64*Q AND q19 is 0.577*Q} \\
                $g'$: \texttt{q1 is 0.657*P AND q20 is 0.609*Q AND q21 is 0.562*P} \\
                $h'$: \texttt{q4 is 0.652*Q AND q20 is 0.679*Q AND q22 is 0.735*P} \\
                $i'$: \texttt{q1 is 0.425*P AND q23 is 0.646*Q} \\
                $j'$: \texttt{q7 is 0.571*Q AND q17 is 0.698*Q} \\
                $k'$: \texttt{q21 is 0.41*P} \\
                $l'$: \texttt{q8 is 0.398*P AND q11 is 0.612*Q AND q12 is 0.644*Q AND q24 is 0.664*P} \\
                $m'$: \texttt{q5 is 0.592*Q AND q15 is 0.617*P AND q24 is 0.402*P} \\
                $n'$: \texttt{q2 is 0.575*Q AND q6 is 0.378*P AND q25 is 0.602*Q} \\
                $o'$: \texttt{q18 is 0.686*Q} \\
                $p'$: \texttt{q22 is 0.342*P AND q25 is 0.607*Q} \\
                $q'$: \texttt{q13 is 0.343*P} \\
                $r'$: \texttt{q2 is 0.682*Q AND q26 is 0.585*Q} \\
                $s'$: \texttt{q14 is 0.651*Q AND q16 is 0.29*P AND q19 is 0.595*Q AND q27 is 0.646*P} \\
                $t'$: \texttt{q9 is 0.336*P AND q23 is 0.681*Q AND q26 is 0.699*Q AND q27 is 0.393*P} \\
                $u'$: \texttt{q10 is 0.393*P}
            };
        \end{tikzpicture}
    }
  \caption{ 
  Left: a coherence graph from our benchmark with edge density target $0.15$ (``sparse''). Center: \texttt{o1-mini} achieved perfect reconstruction on this example under high uncertainty. Right: the modeling propositions incorporated high uncertainty (see \S \ref{sec:uncertainty}).
  }
  \label{fig:example1}
\end{figure}

\begin{figure}[htbp]
  \centering
    \resizebox{.24\textwidth}{!}{%
        % true 2
        \begin{tikzpicture}[
          node distance=2cm, % Adjust as needed
          every node/.style={circle,draw,minimum size=0.7cm,inner sep=0,outer sep=0},
          blue edge/.style={blue, ultra thick},
          red edge/.style={red, ultra thick, dashed},
          xscale=4, % Adjust scaling as needed
          yscale=4,
        ]
        
        % Node positions
        \node (a) at (1.0,1.2957531388962193e-08) {$a$};
        \node (g) at (-0.06824243866721937,0.9976687496274143) {$g$};
        \node (n) at (-0.9172112817530316,-0.39840106602336356) {$n$};
        \node (u) at (0.6825530549588861,-0.7308360757333034) {$u$};
        \node (h) at (-0.3348795685115096,0.942260927854278) {$h$};
        \node (b) at (0.9629172685164649,0.2697967999020556) {$b$};
        \node (j) at (-0.7757112268798297,0.6310879676815739) {$j$};
        \node (p) at (-0.5766802328003474,-0.8169699724797971) {$p$};
        \node (c) at (0.8544194111719118,0.5195839500972502) {$c$};
        \node (d) at (0.6825531145635304,0.7308359824390775) {$d$};
        \node (w) at (0.9629173281211093,-0.2697965355684152) {$w$};
        \node (l) at (-0.9906859268916305,0.13616653994265754) {$l$};
        \node (e) at (0.4600650405102123,0.8878852201033856) {$e$};
        \node (o) at (-0.7757113460891185,-0.6310878821618667) {$o$};
        \node (i) at (-0.5766803520096362,0.8169698791855711) {$i$};
        \node (f) at (0.20345604935770417,0.9790840811114979) {$f$};
        \node (m) at (-0.9906859268916305,-0.13616669284152794) {$m$};
        \node (q) at (-0.33487953870918735,-0.9422609019392152) {$q$};
        \node (k) at (-0.9172112817530316,0.39840115154307076) {$k$};
        \node (t) at (0.46006495110324575,-0.8878852537929671) {$t$};
        \node (r) at (-0.06824237906257498,-0.9976687237123514) {$r$};
        \node (v) at (0.854419291962623,-0.5195841029961206) {$v$};
        \node (s) at (0.20345598975305978,-0.9790840551964352) {$s$};
        
        % Edges
        \draw[red edge] (a) -- (g);
        \draw[red edge] (a) -- (n);
        \draw[blue edge] (a) -- (u);
        \draw[blue edge] (a) -- (h);
        \draw[blue edge] (a) -- (w);
        \draw[red edge] (g) -- (b);
        \draw[red edge] (g) -- (c);
        \draw[red edge] (g) -- (d);
        \draw[red edge] (g) -- (l);
        \draw[blue edge] (n) -- (c);
        \draw[red edge] (n) -- (i);
        \draw[blue edge] (n) -- (s);
        \draw[red edge] (h) -- (k);
        \draw[red edge] (h) -- (q);
        \draw[red edge] (b) -- (j);
        \draw[red edge] (b) -- (p);
        \draw[red edge] (b) -- (c);
        \draw[blue edge] (j) -- (r);
        \draw[blue edge] (j) -- (q);
        \draw[blue edge] (p) -- (c);
        \draw[red edge] (c) -- (d);
        \draw[red edge] (c) -- (w);
        \draw[red edge] (d) -- (l);
        \draw[blue edge] (w) -- (o);
        \draw[blue edge] (w) -- (v);
        \draw[blue edge] (l) -- (i);
        \draw[red edge] (e) -- (o);
        \draw[red edge] (e) -- (i);
        \draw[blue edge] (e) -- (m);
        \draw[blue edge] (o) -- (r);
        \draw[blue edge] (i) -- (f);
        \draw[blue edge] (i) -- (t);
        \draw[blue edge] (f) -- (m);
        \draw[red edge] (f) -- (q);
        \draw[blue edge] (k) -- (t);
        \draw[blue edge] (t) -- (s);
        \draw[blue edge] (r) -- (v);
        
        \end{tikzpicture}
    }
    \resizebox{.24\textwidth}{!}{%
        % reconstruction 2
        \begin{tikzpicture}[
          node distance=2cm, % Adjust as needed
          every node/.style={circle,draw,minimum size=0.7cm,inner sep=0,outer sep=0},
          blue edge/.style={blue, ultra thick},
          red edge/.style={red, ultra thick, dashed},
          xscale=4, % Adjust scaling as needed
          yscale=4,
        ]
        
        % Node positions
        \node (a) at (1.0,1.2957531388962193e-08) {$a$};
        \node (h) at (-0.3348795685115096,0.942260927854278) {$h$};
        \node (u) at (0.6825530549588861,-0.7308360757333034) {$u$};
        \node (w) at (0.9629173281211093,-0.2697965355684152) {$w$};
        \node (e) at (0.4600650405102123,0.8878852201033856) {$e$};
        \node (m) at (-0.9906859268916305,-0.13616669284152794) {$m$};
        \node (f) at (0.20345604935770417,0.9790840811114979) {$f$};
        \node (i) at (-0.5766803520096362,0.8169698791855711) {$i$};
        \node (l) at (-0.9906859268916305,0.13616653994265754) {$l$};
        \node (t) at (0.46006495110324575,-0.8878852537929671) {$t$};
        \node (j) at (-0.7757112268798297,0.6310879676815739) {$j$};
        \node (q) at (-0.33487953870918735,-0.9422609019392152) {$q$};
        \node (r) at (-0.06824237906257498,-0.9976687237123514) {$r$};
        \node (k) at (-0.9172112817530316,0.39840115154307076) {$k$};
        \node (o) at (-0.7757113460891185,-0.6310878821618667) {$o$};
        \node (v) at (0.854419291962623,-0.5195841029961206) {$v$};
        \node (s) at (0.20345598975305978,-0.9790840551964352) {$s$};
        \node (d) at (0.6825531145635304,0.7308359824390775) {$d$};
        \node (g) at (-0.06824243866721937,0.9976687496274143) {$g$};
        \node (n) at (-0.9172112817530316,-0.39840106602336356) {$n$};
        \node (b) at (0.9629172685164649,0.2697967999020556) {$b$};
        \node (c) at (0.8544194111719118,0.5195839500972502) {$c$};
        \node (p) at (-0.5766802328003474,-0.8169699724797971) {$p$};
        
        % Edges
        \draw[blue edge] (a) -- (h);
        \draw[blue edge] (a) -- (u);
        \draw[blue edge] (a) -- (w);
        \draw[red edge] (a) -- (d);
        \draw[red edge] (a) -- (g);
        \draw[red edge] (a) -- (n);
        \draw[red edge] (h) -- (k);
        \draw[red edge] (h) -- (q);
        \draw[red edge] (w) -- (c);
        \draw[blue edge] (e) -- (m);
        \draw[red edge] (e) -- (i);
        \draw[red edge] (e) -- (o);
        \draw[blue edge] (m) -- (f);
        \draw[blue edge] (f) -- (i);
        \draw[red edge] (f) -- (n);
        \draw[red edge] (f) -- (q);
        \draw[blue edge] (i) -- (l);
        \draw[blue edge] (i) -- (t);
        \draw[red edge] (i) -- (n);
        \draw[red edge] (l) -- (d);
        \draw[red edge] (l) -- (g);
        \draw[blue edge] (t) -- (k);
        \draw[blue edge] (t) -- (s);
        \draw[blue edge] (j) -- (q);
        \draw[blue edge] (j) -- (r);
        \draw[red edge] (j) -- (b);
        \draw[blue edge] (r) -- (o);
        \draw[blue edge] (r) -- (v);
        \draw[red edge] (d) -- (c);
        \draw[red edge] (b) -- (c);
        \draw[red edge] (b) -- (p);
        \draw[red edge] (c) -- (p);
        
        \end{tikzpicture}
    }
    \resizebox{.49\textwidth}{!}{%
        % true 2
        \begin{tikzpicture}[
          xscale=4, % Adjust scaling as needed
          yscale=4,
        ]
            \node[align=left] (a) at (0,0) {
                $a'$: \texttt{q1 is 0.344*P AND q2 is 0.611*P AND q3 is 0.595*Q AND q4 is 0.574*Q AND q5 is 0.548*Q} \\
                $b'$: \texttt{q11 is 0.669*P AND q30 is 0.605*Q AND q32 is 0.655*P} \\
                $c'$: \texttt{q6 is 0.625*Q AND q14 is 0.576*P AND q15 is 0.667*P AND q30 is 0.311*Q AND q32 is 0.36*P} \\
                $d'$: \texttt{q14 is 0.366*P AND q31 is 0.664*P} \\
                $e'$: \texttt{q19 is 0.62*P AND q20 is 0.35*P AND q21 is 0.529*Q} \\
                $f'$: \texttt{q23 is 0.509*Q AND q25 is 0.622*Q AND q26 is 0.294*P} \\
                $g'$: \texttt{q1 is 0.655*P AND q30 is 0.452*P AND q31 is 0.633*P} \\
                $h'$: \texttt{q4 is 0.673*Q AND q9 is 0.583*P AND q10 is 0.595*P} \\
                $i'$: \texttt{q7 is 0.429*P AND q18 is 0.606*Q AND q20 is 0.627*P AND q23 is 0.58*Q AND q24 is 0.557*Q} \\
                $j'$: \texttt{q11 is 0.338*P AND q12 is 0.663*Q AND q13 is 0.623*Q} \\
                $k'$: \texttt{q10 is 0.39*P AND q27 is 0.575*Q} \\
                $l'$: \texttt{q18 is 0.729*Q AND q31 is 0.351*P} \\
                $m'$: \texttt{q21 is 0.625*Q AND q25 is 0.596*Q} \\
                $n'$: \texttt{q2 is 0.395*P AND q6 is 0.575*Q AND q7 is 0.707*P AND q8 is 0.616*Q} \\
                $o'$: \texttt{q16 is 0.568*Q AND q19 is 0.298*P AND q22 is 0.656*Q} \\
                $p'$: \texttt{q32 is 0.415*P} \\
                $q'$: \texttt{q9 is 0.286*P AND q12 is 0.531*Q AND q26 is 0.601*P} \\
                $r'$: \texttt{q13 is 0.638*Q AND q22 is 0.585*Q AND q29 is 0.564*Q} \\
                $s'$: \texttt{q8 is 0.658*Q AND q28 is 0.62*Q} \\
                $t'$: \texttt{q24 is 0.628*Q AND q27 is 0.638*Q AND q28 is 0.593*Q} \\
                $u'$: \texttt{q3 is 0.675*Q} \\
                $v'$: \texttt{q17 is 0.656*Q AND q29 is 0.557*Q} \\
                $w'$: \texttt{q5 is 0.612*Q AND q15 is 0.383*P AND q16 is 0.615*Q AND q17 is 0.699*Q}
            };
        \end{tikzpicture}
    }
  \caption{
  Left: another coherence graph from our benchmark with edge density target $0.15$ (``sparse''). Center: \texttt{o1-mini} achieved good but imperfect ($F_1 = 0.96$: see vertices $c$, $d$, $g$, $n$, and $o$) reconstruction on this example under high uncertainty. Right: the modeling propositions incorporated high uncertainty (see \S \ref{sec:uncertainty}).
}
  \label{fig:example2}
\end{figure}

% Applying the proposition generation algorithm described above 
This yields proposition sets involving variable and property counts as detailed in \S \ref{sec:graph_characteristics}. For each problem, we generate four additional sets of propositions. Three of these correspond to the noise regimes described in \S \ref{sec:fuzzy_uncertainty}. We also generate a set of propositions where we add no noise (i.e., $p \leftarrow 1 * p$ and $\lnot p \leftarrow 0 * P$) in order to control for the presence of distractors in the evaluation.

Figure \ref{fig:example1} shows a problem from the benchmark including a synthetic coherence graph with 21 vertices, modeling propositions, and \texttt{o1-mini}'s perfect graph reconstruction (which occurs half the time, as Figure \ref{fig:cohere_graphs} shows). Figure \ref{fig:example2} is similar, but with imperfect reconstruction.

\subsection{\label{sec:prompts}Prompts}
We use a common prompt structure shown in \S \ref{sec:prompt_example} that a) establishes the consistency reasoning task, b) describes variables and properties relevant for evaluating a set of propositions, c) sets fuzzy membership thresholds for properties (see \S \ref{sec:fuzzy_uncertainty}) and d) embeds propositions for a given problem.

\section{\label{sec:results}Results}

\subsection{\label{sec:inference_results}LLMs successfully infer synthetic coherence graphs from a single prompt} 
\begin{figure}[htbp]
    \centering
    \includegraphics[width=1\linewidth, trim={0 8 0 0mm}, clip]{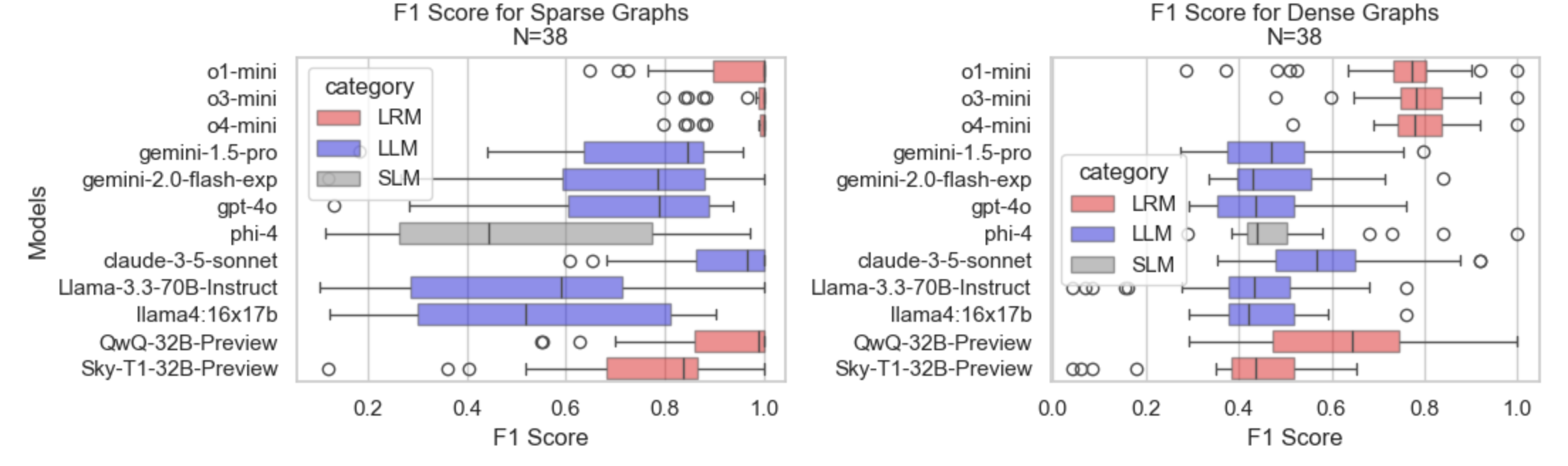}
        \caption{\texttt{o1/3/4-mini}, \texttt{claude-3.5-sonnet}, and \texttt{QwQ-32B} have high micro $F_1$ scores.}
    \label{fig:cohere_graphs}
\end{figure}

Model performance varies considerably, with \texttt{o1/3/4-mini}, \texttt{claude-3.5-sonnet} and \texttt{QwQ-32B} outperforming the other models for both \texttt{sparse} and \texttt{dense} problems: see Figure \ref{fig:cohere_graphs}. 
 
\subsection{Minor post-processing is sometimes necessary to handle obvious reconstruction errors}
We noted four general types of reconstruction errors. First, some reconstructed graphs do not include a given proposition seen in the prompt. We observed this sporadically in all models. Second, we noticed some instances with Gemini models where the reconstructed graph contains nodes named, for instance, "Proposition(a)" instead of a or "(a)" instead of a. We noticed instances where \texttt{QwQ} incorrectly capitalized proposition names, for instance, substituting "A" for "a". We correct these first three error types in post-processing before evaluation.

Finally, we observed that some models hallucinate extra propositions in the reconstructed graph. Out of 608 inference attempts, \texttt{gemini-2.0-flash-exp} hallucinates on 10 inference attempts, \texttt{gpt-4o} on 5, \texttt{gemini-1.5-pro} on 4, and \texttt{Sky-T1-32B} on 1. We do not correct this type of error with post-processing; we instead eliminate these inference attempts from evaluation for these models since these hallucinations can be easily detected at runtime.

\section{\label{sec:conclusion}Conclusion}

While our main result is a demonstration that some LLMs can reconstruct signed coherence graphs with high fidelity, an ideal is to connect a benchmark with realistic examples. 
\footnote{It appears that \texttt{o3/4-mini} significantly outperform \texttt{o1-mini} in practice on more complex (not larger \emph{per se}) real-world instances. However, our results are still too preliminary to be appropriate to detail here.}
This involves two factors: topology and weights. Regarding topology, some realistic examples we have examined are compatible with the ER hypothesis, and others are not. However, our basic approach is compatible with any signed graph topology model (including empirical results from coherence graph construction itself). It is important to note also that in the present context, topology also refers to the structure of cliques and star forest decompositions, both of which admit significant variation to be considered in future work along with underlying graph characteristics beyond uniform sparsity. Regarding weights, extensions of our benchmark to incorporate weights in $[-1,1]$ by (e.g.) building on \S \ref{sec:uncertainty} and a weighted stochastic block model \citep{aicher2015learning} are left for future work. 

Other future extensions of the work could include benchmarking more recent models and variants; deeper analyses along the lines of our appendices, and alternative prompting strategies (e.g., block-decomposing the space of proposition pairs) that are relevant for scaling. 

CDI provides a good model for many forms of cognition, including perception and planning. Combining CDI with consistency evaluations by neural models may lead to advancements in the state of the art in these tasks. Moreover, our results indicate that it is now feasible to computationally instantiate a coherence theory of truth \citep{sep-truth-coherence}. By hard-coding conclusively established propositions, this theory can be anchored in a correspondence theory of truth \citep{sep-truth-correspondence}.

% \section*{Acknowledgments}
% Both authors thank Tejas Patel and Jim Simpson for useful exchanges, and Alexander Medeiros and Alberto Tolentino for software engineering and infrastructure support. SH also thanks George Cybenko, Neil Gerr, Ludmilla Huntsman, Rob Johnston, Michael Robinson, John Santini, Benjamin Wittes, and Levent Yilmaz for useful exchanges. 
% This research was developed with funding from the Defense Advanced Research Projects Agency (DARPA). The views, opinions and/or findings expressed are those of the author(s) and should not be interpreted as representing the official views or policies of the Department of Defense or the U.S. Government. Distribution Statement “A” (Approved for Public Release, Distribution Unlimited).

% \section*{Ethics Statement}
% \steve{No killer robots were harmed in the course of this research.}

\bibliography{benchmarkingCoherenceBib}

\appendix

\section{\label{sec:computation}Coherence can be computed in different ways}

\subsection{\label{sec:computationClassical}Classical coherence is equivalent to 2-MAX-XORSAT and to sparse approximation for 2-XORSAT}

There are a number of algorithms for computing coherence classically. Besides a brute force solution to the MAX-CUT problem for a coherence graph $G_\sigma$, there are also greedy stochastic, simulated annealing, and semidefinite programming approximations \citep{thagard1998coherence,gartner2012approximation}. Historically, the dominant approach is a dynamical neural (``connectionist'') algorithm discussed in \citet{thagard2002coherence}. However, implementations are sufficiently outdated or hard to work with 
\footnote{See, e.g., \url{https://github.com/mars0i/popco}, \url{https://github.com/russellcameronthomas/JavaECHO_command_line}, \url{https://github.com/B-Sabev/ComputationalModelsOfCoherence}, \url{https://github.com/tjd/echo}, or \url{https://github.com/MaxRae/ConnectionistSudoku}. Even the two most recent of these have not been updated for six years as of this writing.}
that an adaptation based on an Ising model has recently been developed \citep{maier2023comparing}. 

We proceed to outline two computational perspectives on coherence that to our knowledge have not previously been considered explicitly, though the second was discussed at the level of prose in \citet{huntsman2024prospects}. Given a coherence graph $G_\sigma$, the consistency equation $\sigma(v,w) = 1$ corresponds to an equation of the form $x_v + x_w = 0$ over the Boolean field $\mathbb{F}_2$, and also to the propositional clause $v \iff w$. Meanwhile, the inconsistency equation $\sigma(v,w) = -1$ corresponds to $x_v + x_w = 1$, and also to the propositional clause $v \oplus w$, where $\oplus = \text{XOR}$. Therefore, we can repackage $G_\sigma$ into the linear equation $Bx = c$, where $B$ is the incidence matrix of $G_\sigma$, or equivalently the biadjacency matrix of the factor graph of the 2-XORSAT/2-Affine-SAT \citep{roy2006fault,mezard2009information} formula obtained from the clauses indicated just above. 

For example, the coherence graph in Figure \ref{fig:variables} corresponds to the linear equation
\begin{equation}
\begin{pmatrix}
1 & 1 & 0 & 0 & 0 \\
0 & 1 & 1 & 0 & 0 \\
0 & 1 & 0 & 1 & 0 \\
0 & 1 & 0 & 0 & 1 \\
0 & 0 & 1 & 1 & 0 \\
0 & 0 & 0 & 1 & 1 \\
1 & 0 & 0 & 0 & 1
\end{pmatrix} 
\begin{pmatrix}
x_a \\ x_b \\ x_c \\ x_d \\ x_e
\end{pmatrix} 
=
\begin{pmatrix}
1 \\ 0 \\ 1 \\ 0 \\ 0 \\ 1 \\ 0
\end{pmatrix} 
\nonumber
\end{equation}
and also to the propositional formula
$$(a \oplus b) \land (b \iff b) \land (b \oplus d) \land (b \iff e) \land (c \iff d) \land (d \oplus e) \land (e \iff a).$$
By using the identities $v \iff w \equiv (\lnot v \lor w) \land (v \lor \lnot w)$ and $v \oplus w \equiv (v \lor w) \land (\lnot v \lor \lnot w)$, we can transform this into a CNF-SAT formula.

Of course, in general the linear equation has no solution and the propositional formula is unsatisfiable. In the propositional context, the coherence computation corresponds to the MAX-2-XORSAT problem, which unsurprisingly reduces to MAX-CUT \citep{moore2011nature}. To our knowledge, the linear algebra formulation of coherence does not appear in the literature. 
\footnote{However, this formulation of the MAX-XORSAT problem is found in the literature on low-density generator matrix codes, where it is equivalent to performing source encoding \citep{wainwright2010lossy}.
}
The goal is to find a certain approximate solution to $Bx = c$, where again $B$ is the incidence matrix (over $\mathbb{F}_2$) of $G_\sigma$, $x$ is a vector of propositional truth values, and $c$ is a vector encoding $\sigma$. Specifically, we want to find the assignment $x$ that satisfies the most rows of the equation: this is equivalent to satisfying the most clauses, i.e., to solving MAX-(2-XOR)SAT all over again. In the language of sparse approximation, we want to solve $\arg \min_x \|Bx-c\|_0$ over $\mathbb{F}_2$, where the $L^0$ ``norm'' (i.e., the size of the support) is indicated.
\footnote{Note that i) over $\mathbb{F}_2$, the $L^0$ and $L^1$ norms are the same, but this offers no advantage; and ii) the usual problem of sparse approximation is of the form $\min_y \|y\|_0$ such that $b = Ay$. Our problem is not expressible in this way and it does not appear to be treated in the literature from the perspective of sparse approximation/optimization, cf. \citep{natarajan1995sparse}. 
}
We can write this as a more conventional-looking optimization with objective $\sum_j (-1)^{c_j + e_j^T B x}$ \citep{jordan2024optimization}. However, there is a more general formulation as an integer linear program \citep{vazirani2001approximation} whose relaxation is of substantial practical interest, as we shall discuss in the sequel.

\subsection{\label{sec:computationGeneral}A general approach: weighted MAX-SAT}

We sketch how to represent higher-order (in)consistency relationships such as trilemmas. A conceptually simple but computationally involved general approach is as follows:
\begin{itemize}
    \item for each ordered pair of dependent claims, rate the consistency of the second given the first, and associate this to a weight for an implication term;
    \item for each ordered triple of dependent claims, rate the consistency of the second and third given the first, and separately the consistency of the third given the first and second: associate these with corresponding implication terms;
    \item etc., truncating this hierarchy as desired/practical. 
\end{itemize}
Note that this requires a clique-finding algorithm such as that of \citet{chiba1985arboricity}, and that each unordered tuple corresponds to several ordered tuples, ensuring symmetry.

While this approach appears to adequately generalize the notion of a coherence graph into what amounts to a weighted \emph{simplicial complex}---i.e., a weighted hypergraph containing all sub-hyperedges \citep{rosiak2022sheaf}---it is not yet suitable for practical MAX-SAT solvers of any sort. These only address weighted CNF-SAT problems, while common translations into CNF-SAT do not preserve weights correctly. Therefore, applying a MAX-SAT solver would generally require (implementing and) applying a specialized transformation of the sort detailed in \citet{li2022clausal}, which to our knowledge has no public implementation.

Once a coherence problem is properly transformed into CNF-SAT, there is an interesting alternative to weighted MAX-SAT solvers. An efficient probabilistic algorithm for weighted MAX-SAT discussed in \cite{vazirani2001approximation} provides a natural probabilistic interpretation that is certain to be useful in many if not most practical situations. For example, a proposition with acceptance probability near 0.5 may indicate a need for---and even drive the collection of---additional data for incorporation into the relevant structure.

The general approach to representing and solving ``higher-order'' coherence problems outlined just above was suggested by \citet{huntsman2024prospects}, who were unaware of prior work on coherence \emph{per se}. The underlying motivation was that a \emph{sheaf} ought to represent consistency data. Briefly, a sheaf is comprised of data on open sets in a topological space that satisfy natural restriction---equivalently, gluing---axioms \citep{rosiak2022sheaf}. The connection to sheaves is that every logical clause in a CNF-SAT formula introduces a logical constraint that must be satisfied in order for the overall formula to be satisfied. In particular, adding clauses can never enlarge the solution space \citep{srinivas1993applications}.

It is not immediately clear how to adapt our approach for modeling coherence graphs to this more general setting. We leave this for future work.

\section{\label{sec:vincennes}A case study comparing human and LLM consistency ratings}

There is additional evidence that LLMs can reliably gauge propositional (in)consistency \citep{huntsman2024prospects,kumar2024nlp}. However, we are not aware of any assessment of performance relative to humans. Here, we outline a case study performed in summer 2024 in which \texttt{ChatGPT 4} demonstrated superhuman performance at gauging (in)consistency.

In \citet{thagard1992adversarial}, a coherence graph modeled a decision facing the captain of the cruiser USS \emph{Vincennes} on 3 July 1988: was radar track 4131 taking off from the Iranian civilian-military airfield at Bandar Abbas a civilian aircraft, or a hostile F-14 preparing to attack? The coherence graph included ``positive evidence'' propositions (labeled E*) copied almost verbatim from \S III.C.1.b of the investigation report \citep{fogarty1988formal} along with a few of their negations (labeled NE*) and hypothetical propositions that the track was attacking (labeled A*) or commercial (labeled C*). As is still common today for analyses using CDI, the edge set and weights (here, simply signs) were constructed by hand.

\begin{figure}[htbp]
    \centering
    \includegraphics[width=.45\linewidth, trim={110 55 95 20mm}, clip]{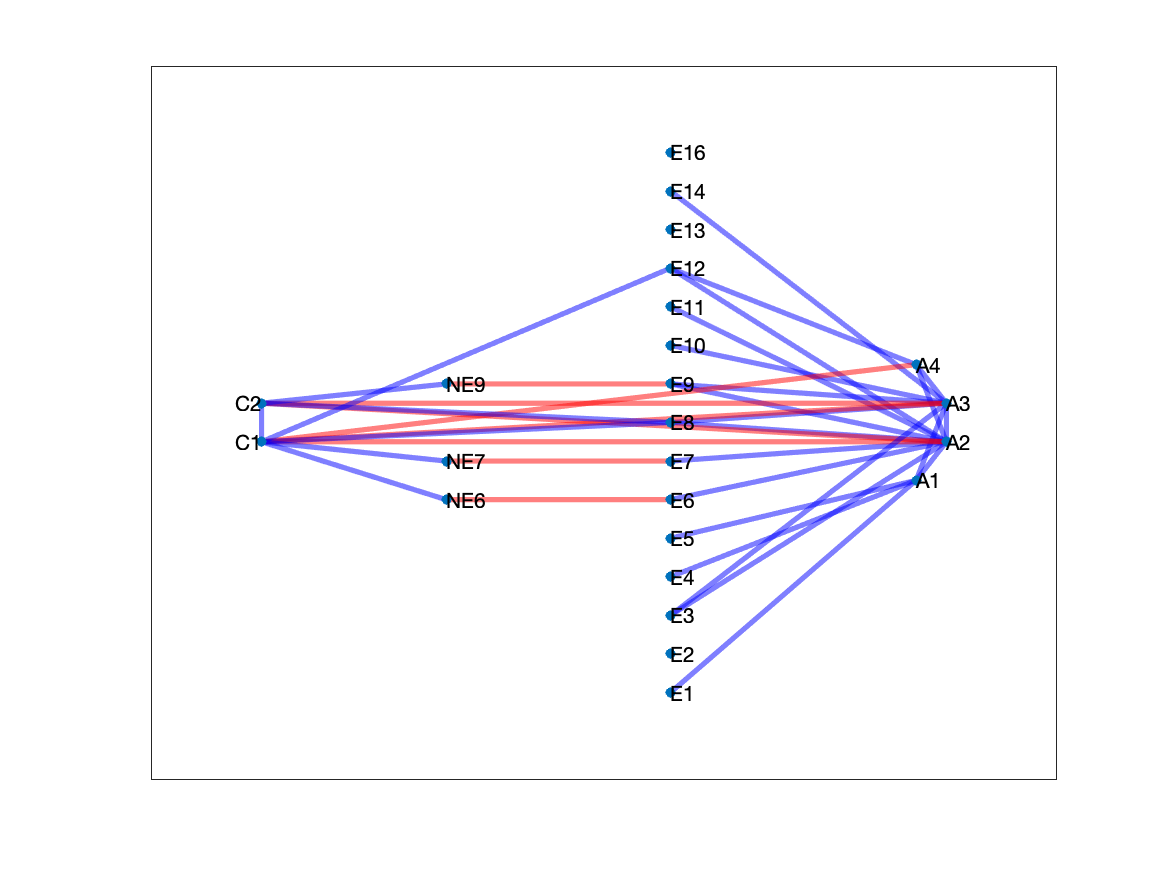}
    \quad \quad
    \includegraphics[width=.45\linewidth, trim={110 55 95 20mm}, clip]{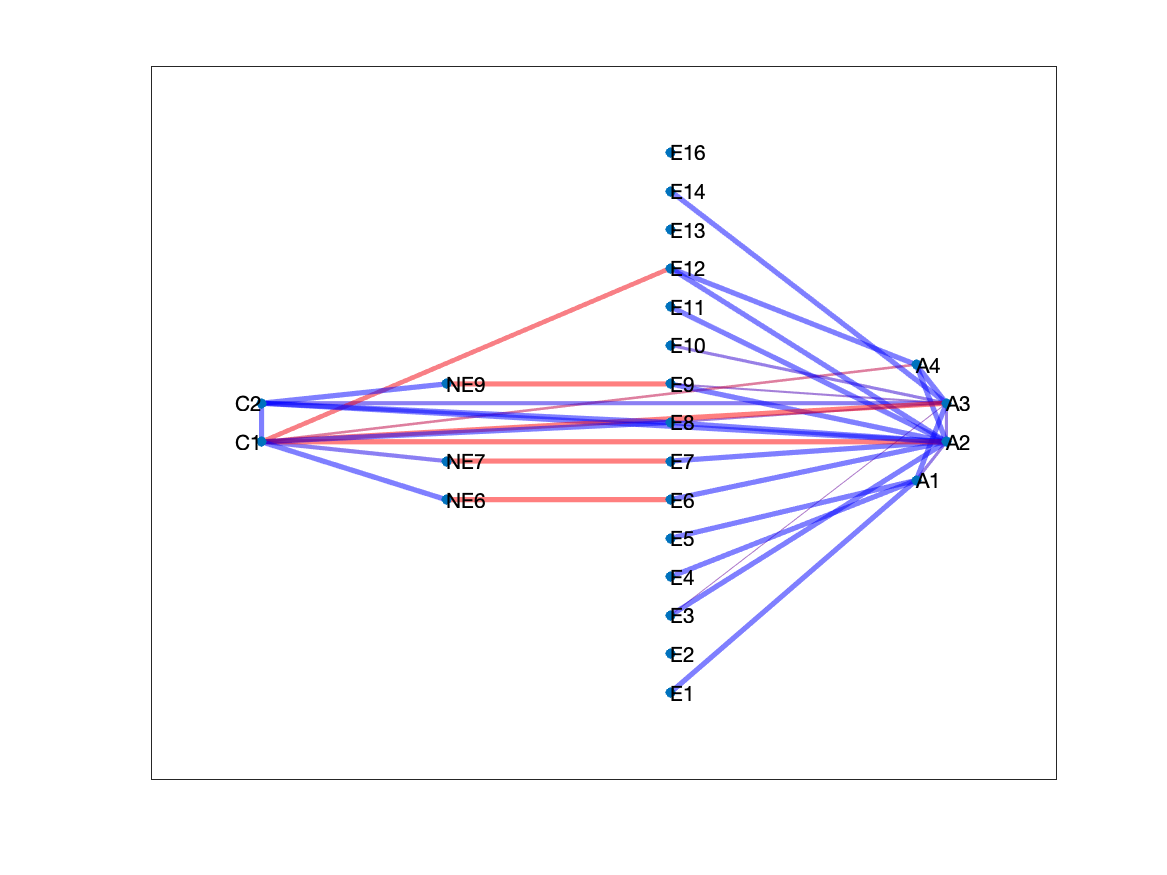}
    \caption{Left: A hand-crafted coherence graph from \citet{thagard1992adversarial}. {\color{blue}Blue} and {\color{red}red} respectively indicate {\color{blue}positive} and {\color{red}negative} edge weights. Right: A coherence graph with the same edges, but with weights given by rescaled median consistency ratings from \texttt{ChatGPT 4}. Thickness indicates the relative magnitude of weights.
}
    \label{fig:vincennesGraphs}
\end{figure}

\begin{figure}[htbp]
    \centering
    \includegraphics[width=1\linewidth]{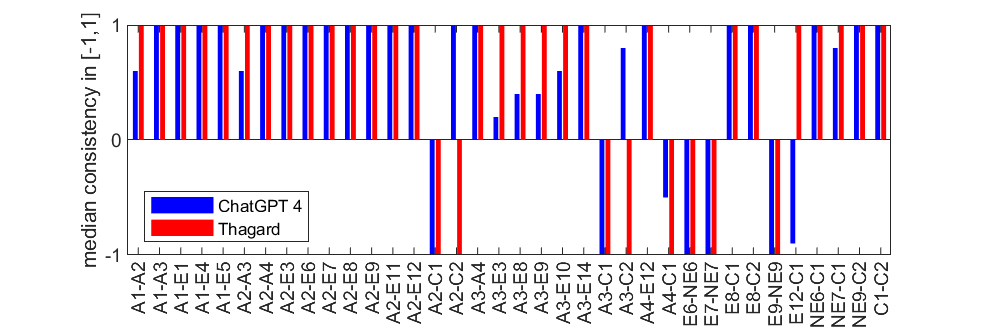}
    \caption{{\color{blue}Rescaled median consistency ratings by \texttt{ChatGPT 4}} \emph{versus} {\color{red}hand-crafted consistency ratings from \citet{thagard1992adversarial}}.
}
    \label{fig:vincennesBars}
\end{figure}

To focus on \texttt{ChatGPT 4}'s performance in evaluating (in)consistency, we considered the same edge set as the manually-constructed coherence graph, i.e., we neglected any considerations of dependence.
\begin{quote}
    It is important to note here that (the nontrivial connected component of) the coherence graph has $|V| = 21$ and $|E| = 36$, so its edge density is $36/210 = 0.17$. {\bf In other words, this example is in exactly the same regime as the larger and sparser benchmark graphs discussed in the main text.}
\end{quote}
For each edge, we asked \texttt{ChatGPT 4} to rate the consistency of the corresponding pairs of propositions 10 times using a variant of the prompt in \citet{huntsman2024prospects} that included an extract from the executive summary of \citet{fogarty1988formal} as background information. The results are shown in Figures \ref{fig:vincennesGraphs} and \ref{fig:vincennesBars}. Supporting data and scripts are in \citet{huntsman2025vincennes}.

A bit of introspection and common sense reveals that for the largest discrepancies between the median LLM rating and the human rating, the median LLM rating is more reasonable in each event. Consider the four largest discrepancies, listed in order of their appearance in Figure \ref{fig:vincennesBars}:
\begin{itemize}
    \setlength{\itemindent}{.25in}
    \item[A2-C2:] These are obviously consistent, so \texttt{ChatGPT 4}'s rating is better than the rating in \citet{thagard1992adversarial}.
    \begin{itemize}
        \item A2 is ``Track 4131 was an F-14.''
        \item C2 is ``Track 4131 was taking off.''
    \end{itemize}
    \item[A3-E3:] Examination of outputs shows that \texttt{ChatGPT 4} considered plausible technical failures and misunderstandings. In light of this, it is fair to conclude that \texttt{ChatGPT 4}'s rating is better than the rating in \citet{thagard1992adversarial}.
    \begin{itemize}
        \item A3 is ``Track 4131 intended to attack.''
        \item C3 is ``Track 4131 was not responding to verbal warnings over [air distress frequencies].''
    \end{itemize}
    \item[A3-C2:] Both A3 and C2 are listed above. These are obviously consistent, so \texttt{ChatGPT 4}'s rating is better than the rating in \citet{thagard1992adversarial}.
    \item[E12-C1:] Examination of outputs shows that \texttt{ChatGPT 4} considered navigation and communications emissions of commercial airliners. In light of this, it is fair to conclude that \texttt{ChatGPT 4}'s rating is better than the rating in \citet{thagard1992adversarial}.
    \begin{itemize}
        \item E12 is ``No [electronic emissions were reported] from track 4131, however, F-14s can fly [without electronic emissions].''
        \item C1 is ``Track 4131 was a commercial airliner.''
    \end{itemize}
\end{itemize}

To summarize, \texttt{ChatGPT 4} demonstrated superhuman performance at gauging the (in)consistency of propositions.

\section{\label{sec:anova} ANOVA for micro $F_1$}

The two-way ANOVA for micro $F_1$ in Table \ref{tab:anova} shows significant effects.
% Requires: \usepackage{array}

\begin{table}[h]
    \centering
    % \begin{tabular}{lrrrr}
    %     \toprule
    %     & \textbf{SS} & \textbf{DF} & \textbf{F} & $\textbf{p-value}$ \\
    %     \midrule
    %     \textbf{Model} & 9.69 & 8.0 & 29.62 & $<0.001$  \\
    %     \textbf{Sparsity} & 9.35 & 1.0 &228.51 & $<0.001$  \\
    %     \textbf{Model $\times$ sparsity} & 1.30 & 8.0 & 3.97 & $<0.001$  \\
    %     \textbf{Error} & 27.77 & 679.0 & \texttt{NaN} & \texttt{NaN} \\
    %     \bottomrule
    % \end{tabular}
    \begin{tabular}{lrrrr}
        \toprule
        & \textbf{SS} & \textbf{DF} & \textbf{F} & $\textbf{p-value}$ \\
        \midrule
        \textbf{C(model)} & 17.86 & 11.0 & 60.77 & $<0.001$ \\
        \textbf{density} & 9.87 & 1.0 & 369.50 & $<0.001$ \\
        \textbf{C(model):density} & 1.84 & 11.0 & 6.28 & $<0.001$ \\
        \textbf{residual} & 22.87 & 856.0 & \texttt{NaN} & \texttt{NaN} \\
        \bottomrule
    \end{tabular}
    \caption{A two-way ANOVA shows significance for model, sparsity, and their interaction.}
    \label{tab:anova}
\end{table}

\section{\label{sec:uncertainty}Modeling and reconstruction under uncertainty}

\subsection{\label{sec:fuzzy_uncertainty} Syntax can model uncertainty in propositions}

To evaluate model judgments of consistency under uncertainty, we borrow from \citet{Ragin2006}. We introduce uncertainty to $p$ or $\lnot p$ by sampling from triangular fuzzy membership functions defined over the unit interval, with $1$ and $0$ respectively corresponding to $p$ and $\lnot p$. After adding uncertainty, a given $p$-assignment is syntactically represented as something like $0.619*p$, while a given $\lnot p$ assignment is syntactically represented as something like $0.346*p$. 
That is, we indicate uncertainty as in the right panels of Figures \ref{fig:example1} and \ref{fig:example2}, with results in Table \ref{tab:graph_uncertainty_01312025}: meanwhile, the prompt in \S \ref{sec:prompt_example} echoes this construction.
With $p \leftarrow \alpha*p$ and $\lnot p \leftarrow \alpha_\lnot*p$ we delineate three uncertainty regimes. The \texttt{low} regime has $\alpha \in [0.75, 1]$ and $\alpha_\lnot \in [0, 0.25]$; the \texttt{medium} regime has $\alpha \in [0.625,0.75]$ and $\alpha_\lnot \in [0.25, 0.375]$, and the \texttt{high} regime has $\alpha \in [0.5, 0.625]$ and $\alpha_\lnot \in [0.375, 0.5]$. % Note that almost surely $\alpha_\lnot \neq 1-\alpha$.

Encoding numerical uncertainty in natural language in a way that ensures accurate interpretation is complicated by individual differences in interpretation related to terms such as ``approximately,'' ``certainly,'' ``about,'' ``exactly,'' etc. \citep{Krisper2019, Ferson2015}. This presents a challenge for, e.g., communicating risks related to climate change, risks of developing complications from medical treatment, or the likelihood of a given geopolitical event in the future \citep{ICD203}. The fuzzy set method above encodes diverse qualitative statements with a consistent formalism, and is directly relevant for use with words of estimative probability \citep{ICD203, kachynskyi2019national}.

\subsection{\label{sec:uncertain_inference_results}LLMs successfully infer synthetic coherence graphs under uncertainty}
Table \ref{tab:graph_uncertainty_01312025} shows that introducing uncertainty does not degrade graph reconstruction fidelity. 

In Figure \ref{fig:uncertainty_examples}, we show how recursively passing an inferred uncertain coherence graph to \texttt{o1-mini} and prompting it to assign weights reflective of the level of uncertainty in property assignments can yield a weighted coherence graph with appropriate edge weights (less certain assignments yield smaller magnitude weights; more certain assignments yield larger magnitude weights).

\begin{figure}[htbp]
\centering
\vspace{-1cm}
\begin{minipage}{0.45\textwidth}
\centering
\begin{tikzpicture}[
  node distance=1cm,
  every node/.style={circle, draw, minimum size=.5cm, inner sep=0.1cm, align=center},
  xscale=.45, % Adjust scaling as needed
  yscale=.45,
]

  % Nodes in a circle
  \node[draw,circle,fill=white,minimum size=6.5mm] (a) at (0:3) {$a$};
  \node[draw,circle,fill=white,minimum size=6.5mm] (b) at (51.4:3) {$b$};
  \node[draw,circle,fill=white,minimum size=6.5mm] (e) at (102.8:3) {$e$};
  \node[draw,circle,fill=white,minimum size=6.5mm] (d) at (154.2:3) {$d$};
  \node[draw,circle,fill=white,minimum size=6.5mm] (f) at (205.6:3) {$f$};
  \node[draw,circle,fill=white,minimum size=6.5mm] (c) at (257:3) {$c$};
  \node[draw,circle,fill=white,minimum size=6.5mm] (g) at (308.4:3) {$g$};

  % Edges with variable color/transparency based on weight
  \draw[dashed, color={rgb,1:red,0.850;green,0.000;blue,0.150}, opacity=0.70, line width=0.7mm] (a) -- (b);
  \draw[solid, color={rgb,1:red,0.150;green,0.000;blue,0.850}, opacity=0.70, line width=0.7mm] (b) -- (e);
  \draw[solid, color={rgb,1:red,0.050;green,0.000;blue,0.950}, opacity=0.90, line width=0.9mm] (e) -- (d);
  \draw[dashed, color={rgb,1:red,0.850;green,0.000;blue,0.150}, opacity=0.70, line width=0.7mm] (e) -- (f);
  \draw[solid, color={rgb,1:red,0.050;green,0.000;blue,0.950}, opacity=0.90, line width=0.9mm] (d) -- (g);
  \draw[solid, color={rgb,1:red,0.150;green,0.000;blue,0.850}, opacity=0.70, line width=0.7mm] (f) -- (c);

  % Propositions list
  \node[align=left] at (9.5, 0) [draw=none, text width=9cm, scale=0.55] {
    \textbf{Propositions (low \textbf{uncertainty}):} \\
    $a$: \texttt{q1 is 0.898*P} \\
    $b$: \texttt{q1 is 0.122*P AND q2 is 0.943*Q} \\
    $c$: \texttt{q5 is 0.887*Q} \\
    $d$: \texttt{q4 is 0.904*Q AND q6 is 0.929*Q} \\
    $e$: \texttt{q2 is 0.874*Q AND q3 is 0.838*P} \\ \hspace*{1cm} \texttt{AND q4 is 0.943*Q} \\
    $f$: \texttt{q3 is 0.085*P AND q5 is 0.962*Q} \\
    $g$: \texttt{q6 is 0.863*Q} \\
  };
\end{tikzpicture}
\end{minipage}
\quad \quad
\begin{minipage}{0.45\textwidth}
\centering
\begin{tikzpicture}[
  node distance=1cm,
  every node/.style={circle, draw, minimum size=.5cm, inner sep=0.1cm, align=center},
  xscale=.45, % Adjust scaling as needed
  yscale=.45,
]

  % Nodes in a circle
  \node[draw,circle,fill=white,minimum size=6.5mm] (a) at (0:3) {$a$};
  \node[draw,circle,fill=white,minimum size=6.5mm] (b) at (51.4:3) {$b$};
  \node[draw,circle,fill=white,minimum size=6.5mm] (e) at (102.8:3) {$e$};
  \node[draw,circle,fill=white,minimum size=6.5mm] (d) at (154.2:3) {$d$};
  \node[draw,circle,fill=white,minimum size=6.5mm] (f) at (205.6:3) {$f$};
  \node[draw,circle,fill=white,minimum size=6.5mm] (c) at (257:3) {$c$};
  \node[draw,circle,fill=white,minimum size=6.5mm] (g) at (308.4:3) {$g$};

  % Edges with variable color/transparency based on weight
  \draw[dashed, color={rgb,1:red,0.750;green,0.000;blue,0.250}, opacity=0.50, line width=0.5mm] (a) -- (b);
  \draw[solid, color={rgb,1:red,0.200;green,0.000;blue,0.800}, opacity=0.60, line width=0.6mm] (b) -- (e);
  \draw[solid, color={rgb,1:red,0.200;green,0.000;blue,0.800}, opacity=0.60, line width=0.6mm] (e) -- (d);
  \draw[dashed, color={rgb,1:red,0.800;green,0.000;blue,0.200}, opacity=0.60, line width=0.6mm] (e) -- (f);
  \draw[solid, color={rgb,1:red,0.200;green,0.000;blue,0.800}, opacity=0.60, line width=0.6mm] (d) -- (g);
  \draw[solid, color={rgb,1:red,0.200;green,0.000;blue,0.800}, opacity=0.60, line width=0.6mm] (f) -- (c);

  % Propositions list
  \node[align=left] at (9.5, 0) [draw=none, text width=9cm, scale=0.55] {
\textbf{Propositions (high \textbf{uncertainty}):} \\
    $a$: \texttt{q1 is 0.620*P} \\
    $b$: \texttt{q1 is 0.376*P AND q2 is 0.547*Q} \\
    $c$: \texttt{q5 is 0.677*Q} \\
    $d$: \texttt{q4 is 0.534*Q AND q6 is 0.604*Q} \\
    $e$: \texttt{q2 is 0.643*Q AND q3 is 0.667*P} \\ \hspace*{1cm} \texttt{AND q4 is 0.643*Q} \\
    $f$: \texttt{q3 is 0.341*P AND q5 is 0.616*Q} \\
    $g$: \texttt{q6 is 0.541*Q} \\
  };
\end{tikzpicture}
\end{minipage}
\vspace{-1cm}
\caption{\texttt{o1-mini}-inferred weights have larger magnitude (indicated here by thickness as well as by color and opacity) when uncertainty is lower (as in the left panel). Both left and right panels reflect perfect reconstructions of the underlying graph.}
\label{fig:uncertainty_examples}
\end{figure}

\begin{table}[h]
\centering
\small
\begin{tabular}{lllllll}
\toprule
\textbf{model} & \textbf{sparsity} & \textbf{base case} & \textbf{zero unc.} & \textbf{low unc.} & \textbf{med unc.} & \textbf{high unc.} \\
\midrule
{\textbf{o1-mini}} & \textbf{sparse} & 0.94 $\pm$ 0.10 & 0.93 $\pm$ 0.10 & 0.93 $\pm$ 0.10 & 0.93 $\pm$ 0.10 & 0.94 $\pm$ 0.11 \\
\textbf{} & \textbf{dense} & 0.74 $\pm$ 0.14 & 0.71 $\pm$ 0.16 & 0.68 $\pm$ 0.16 & 0.73 $\pm$ 0.16 & 0.75 $\pm$ 0.13 \\
\cline{1-7}
{\textbf{o3-mini}} & \textbf{sparse} & 0.97 $\pm$ 0.06 & 0.97 $\pm$ 0.07 & 0.97 $\pm$ 0.06 & 0.97 $\pm$ 0.06 & 0.97 $\pm$ 0.06 \\
\textbf{} & \textbf{dense} & 0.78 $\pm$ 0.10 & 0.77 $\pm$ 0.09 & 0.79 $\pm$ 0.08 & 0.77 $\pm$ 0.08 & 0.79 $\pm$ 0.07 \\
\cline{1-7}
{\textbf{o4-mini}} & \textbf{sparse} & 0.98 $\pm$ 0.06 & 0.98 $\pm$ 0.06 & 0.98 $\pm$ 0.06 & 0.98 $\pm$ 0.06 & 0.98 $\pm$ 0.06 \\
\textbf{} & \textbf{dense} & 0.79 $\pm$ 0.09 & 0.80 $\pm$ 0.07 & 0.80 $\pm$ 0.07 & 0.80 $\pm$ 0.07 & 0.80 $\pm$ 0.07 \\
\cline{1-7}
{\textbf{Gemini 1.5}} & \textbf{sparse} & 0.74 $\pm$ 0.18 & 0.75 $\pm$ 0.18 & 0.77 $\pm$ 0.15 & 0.81 $\pm$ 0.13 & 0.77 $\pm$ 0.17 \\
\textbf{} & \textbf{dense} & 0.48 $\pm$ 0.14 & 0.48 $\pm$ 0.16 & 0.48 $\pm$ 0.13 & 0.49 $\pm$ 0.14 & 0.48 $\pm$ 0.12 \\
\cline{1-7}
{\textbf{Gemini 2.0}} & \textbf{sparse} & 0.71 $\pm$ 0.23 & 0.68 $\pm$ 0.23 & 0.72 $\pm$ 0.20 & 0.73 $\pm$ 0.19 & 0.73 $\pm$ 0.22 \\
\textbf{} & \textbf{dense} & 0.49 $\pm$ 0.13 & 0.47 $\pm$ 0.11 & 0.46 $\pm$ 0.13 & 0.44 $\pm$ 0.10 & 0.48 $\pm$ 0.12 \\
\cline{1-7}
{\textbf{GPT-4o}} & \textbf{sparse} & 0.73 $\pm$ 0.19 & 0.75 $\pm$ 0.16 & 0.74 $\pm$ 0.21 & 0.74 $\pm$ 0.18 & 0.73 $\pm$ 0.17 \\
\textbf{} & \textbf{dense} & 0.46 $\pm$ 0.14 & 0.43 $\pm$ 0.10 & 0.44 $\pm$ 0.14 & 0.46 $\pm$ 0.15 & 0.44 $\pm$ 0.12 \\
\cline{1-7}
{\textbf{Phi-4}} & \textbf{sparse} & 0.53 $\pm$ 0.29 & 0.43 $\pm$ 0.31 & 0.42 $\pm$ 0.30 & 0.41 $\pm$ 0.28 & 0.43 $\pm$ 0.24 \\
\textbf{} & \textbf{dense} & 0.49 $\pm$ 0.13 & 0.49 $\pm$ 0.16 & 0.47 $\pm$ 0.12 & 0.50 $\pm$ 0.16 & 0.50 $\pm$ 0.13 \\
\cline{1-7}
{\textbf{Claude 3.5}} & \textbf{sparse} & 0.92 $\pm$ 0.11 & 0.91 $\pm$ 0.13 & 0.91 $\pm$ 0.12 & 0.91 $\pm$ 0.12 & 0.91 $\pm$ 0.11 \\
\textbf{} & \textbf{dense} & 0.59 $\pm$ 0.15 & 0.61 $\pm$ 0.15 & 0.59 $\pm$ 0.15 & 0.60 $\pm$ 0.14 & 0.62 $\pm$ 0.15 \\
\cline{1-7}
{\textbf{Llama 3.3}} & \textbf{sparse} & 0.54 $\pm$ 0.26 & 0.53 $\pm$ 0.26 & 0.55 $\pm$ 0.22 & 0.62 $\pm$ 0.23 & 0.50 $\pm$ 0.26 \\
\textbf{} & \textbf{dense} & 0.42 $\pm$ 0.16 & 0.47 $\pm$ 0.15 & 0.43 $\pm$ 0.16 & 0.46 $\pm$ 0.16 & 0.42 $\pm$ 0.15 \\
\cline{1-7}
{\textbf{QwQ}} & \textbf{sparse} & 0.92 $\pm$ 0.13 & 0.90 $\pm$ 0.14 & 0.90 $\pm$ 0.13 & 0.91 $\pm$ 0.13 & 0.91 $\pm$ 0.13 \\
\textbf{} & \textbf{dense} & 0.61 $\pm$ 0.19 & 0.59 $\pm$ 0.18 & 0.59 $\pm$ 0.19 & 0.60 $\pm$ 0.20 & 0.62 $\pm$ 0.21 \\
\cline{1-7}
{\textbf{Sky-T1}} & \textbf{sparse} & 0.76 $\pm$ 0.18 & 0.68 $\pm$ 0.23 & 0.62 $\pm$ 0.28 & 0.67 $\pm$ 0.21 & 0.68 $\pm$ 0.24 \\
\textbf{} & \textbf{dense} & 0.43 $\pm$ 0.14 & 0.43 $\pm$ 0.13 & 0.41 $\pm$ 0.14 & 0.42 $\pm$ 0.13 & 0.43 $\pm$ 0.13 \\
\cline{1-7}
\bottomrule
\end{tabular}
\caption{LLMs successfully infer synthetic coherence graphs under uncertainty (unc).}
\label{tab:graph_uncertainty_01312025}
\end{table}

\section{\label{sec:construction} Benchmark generation parameters can be altered to synthesize problem sets with desired characteristics} 

\begin{figure}[htbp]
    \centering
    \includegraphics[width=1\linewidth]{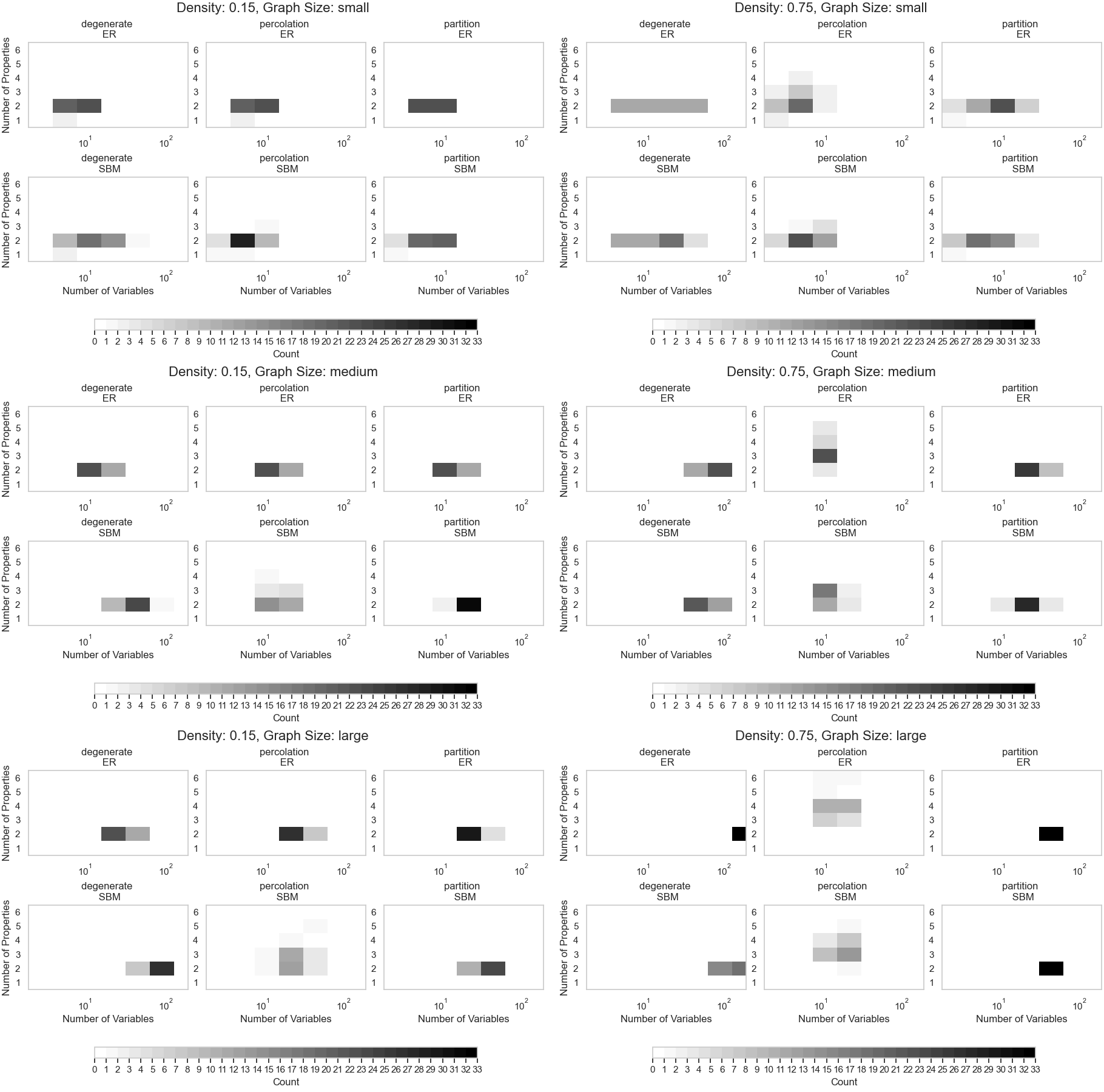}
    \caption{Figure \ref{fig:variables_and_properties} Changing benchmark generation parameters leads to increasing numbers of variables in the resulting synthetic problem sets as graph sizes increase. Top row: small graphs (5-11 propositions). Middle row: medium graphs (13-17 propositions). Bottom row: large graphs (19-23 props). In general, the \texttt{degenerate} edge clique processing method and the \texttt{partition} method (first and third subcolumns in each panel) lead to more variables and fewer properties than the \texttt{percolation} method (middle column of each panel).
} 
    \label{fig:variables_and_properties}
\end{figure}

Figure \ref{fig:variables_and_properties} demonstrates how benchmark generation parameters lead to increasing numbers of variables in the resulting synthetic problem sets as graph sizes increase. The top row shows data for small graphs (5-11 propositions), middle row: medium graphs (13-17 propositions), bottom row: large graphs (19-23 propositions). In general, the \texttt{degenerate} edge clique processing method and the \texttt{partition} method (first and third subcolumns in each panel) lead to more variables and fewer properties than the \texttt{percolation} method (middle column of each panel). 

\section{\label{sec:cross_encoders}Inferring logical entailment with \texttt{ModernBERT} fine-tuned for NLI}

 We performed a preliminary experiment to test whether we could directly identify either a) relevance (defined as two propositions sharing a single variable) or b) consistency directly from white box LLM embeddings or from a number of simple quantifications of inter-token attention in the models' attention mechanisms. While these experiments showed a consistently high error rate (suggesting the need for a more sophisticated mechanistic interpretabilty study along the lines described by \citealt{nanda2023progress}) we report the best results, which 
 came from using a small BERT model (\texttt{ModernBERT} fine-tuned to infer entailment relations between propositions \citep{sileo-2024-tasksource, reimers2019sentencebertsentenceembeddingsusing}. 
 
 Using propositions in the simplified natural language grammar of our benchmark shows accurate entailment inferences along the diagonal (left figure), but multiple errors in off-diagonal relations (e.g., inferring a contradiction in (2,4) between "$q_2$ is !P" and "$q_4$ is !P", among others). However, rewriting propositions more formally (right) shows that \texttt{ModernBERT} is able to perfectly infer consistency relations among these propositions.  

Our tests to extract logical relations directly from attention encodings included estimating e.g., mean or median inter-token attention at the first layer, last layer, or across all layers for \texttt{QwQ-32B} (the most performant open source LLM for our purposes) using benchmark propositions and propositions rewritten in propositional logic. We also tested applying regression to correct positional and autoregressive attention effects. These experiments were inconclusive, suggesting that more complex statistical measures might be needed to identify reasoning mechanisms in white box models.

\begin{figure}[htbp]
    \centering
\includegraphics[width=1\linewidth]{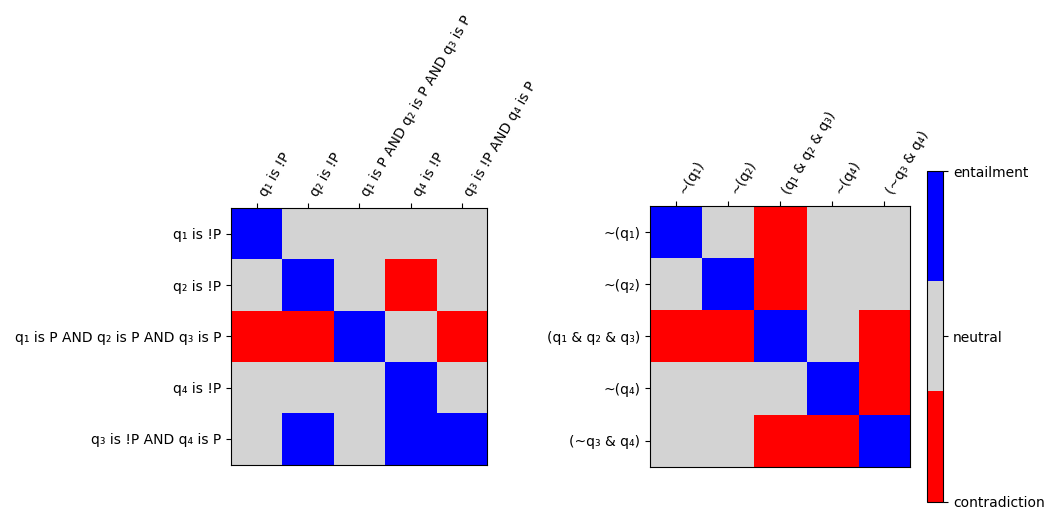}
      \caption{Pairwise natural language inference using \texttt{ModernBERT} for propositions of a simple one property problem shows errors on benchmark propositions (left) but perfect accuracy with propositions transformed to propositional logic (right).}
    \label{fig:v}
\end{figure}

\section{\label{sec:graph_characteristics} Characteristics of synthetic coherence graphs included in our benchmark experiment}

In Table \ref{tab:graph_characteristics}, we show that our experiment includes coherence graphs ranging in size from 5 to 23 propositions. We note that our requirement that all synthetic coherence graphs be connected makes harder to meet the desired sparsity (\texttt{sparse} and \texttt{dense} corresponding to target densities of 0.15 and 0.75 respectively) levels for smaller graphs (5-11 propositions).

\begin{table}[htbp]
    \centering
    \small
    \begin{tabular}{cc|c|c|c|c|c}
        \toprule
        \textbf{n propositions} & \textbf{sparsity} & \textbf{n variables} & \textbf{n properties} & \textbf{edges} & \textbf{density} & \textbf{problem} \\
        & & median & median & median & mean & count \\
        % \hline
        \midrule
        5 & \textbf{sparse} & 4.0 & 1.5 & 4.0 & 0.400000 & 4 \\
          & \textbf{dense} & 3.0 & 2.0 & 7.0 & 0.700000 & 4 \\
        % \hline
        \addlinespace
        7 & \textbf{sparse} & 6.0 & 2.0 & 6.0 & 0.285714 & 4 \\
          & \textbf{dense} & 5.5 & 2.0 & 15.0 & 0.714286 & 4 \\
        % \hline
        \addlinespace
        9 & \textbf{sparse} & 8.0 & 2.0 & 8.0 & 0.222222 & 4 \\
          & \textbf{dense} & 6.5 & 2.0 & 27.0 & 0.750000 & 4 \\
        % \hline
        \addlinespace
        11 & \textbf{sparse} & 10.0 & 2.0 & 10.0 & 0.181818 & 4 \\
           & \textbf{dense} & 9.5 & 3.0 & 41.0 & 0.745455 & 4 \\
        % \hline
        \addlinespace
        13 & \textbf{sparse} & 12.0 & 2.0 & 12.0 & 0.153846 & 4 \\
           & \textbf{dense} & 13.5 & 2.0 & 58.0 & 0.743590 & 4 \\
        % \hline
        \addlinespace
        15 & \textbf{sparse} & 15.0 & 2.0 & 15.0 & 0.142857 & 4 \\
           & \textbf{dense} & 17.5 & 2.5 & 78.0 & 0.742857 & 4 \\
        % \hline
        \addlinespace
        17 & \textbf{sparse} & 19.0 & 2.0 & 20.0 & 0.147059 & 4 \\
           & \textbf{dense} & 23.0 & 2.5 & 102.0 & 0.750000 & 4 \\
        % \hline
        \addlinespace
        19 & \textbf{sparse} & 22.0 & 2.0 & 25.0 & 0.146199 & 4 \\
           & \textbf{dense} & 29.0 & 2.5 & 128.0 & 0.748538 & 4 \\
        % \hline
        \addlinespace
        21 & \textbf{sparse} & 27.0 & 2.0 & 31.0 & 0.147619 & 3 \\
           & \textbf{dense} & 42.0 & 2.0 & 157.0 & 0.747619 & 3 \\
        % \hline
        \addlinespace
        23 & \textbf{sparse} & 32.0 & 2.0 & 37.0 & 0.146245 & 3 \\
           & \textbf{dense} & 52.0 & 2.0 & 189.0 & 0.747036 & 3 \\
        \bottomrule
    \end{tabular}
    \caption{Our experimental benchmark includes 76 graphs (N=38 \texttt{sparse} and N=38 \texttt{dense}) ranging in size from 5 to 23 propositions.
    }
    \label{tab:graph_characteristics}
\end{table}

\section{\label{sec:l1_distances} $L^1$/graph edit distance convergence}

\begin{figure}[htbp]
    \centering
    \includegraphics[width=1\linewidth, trim={0 0 0 0mm}, clip]{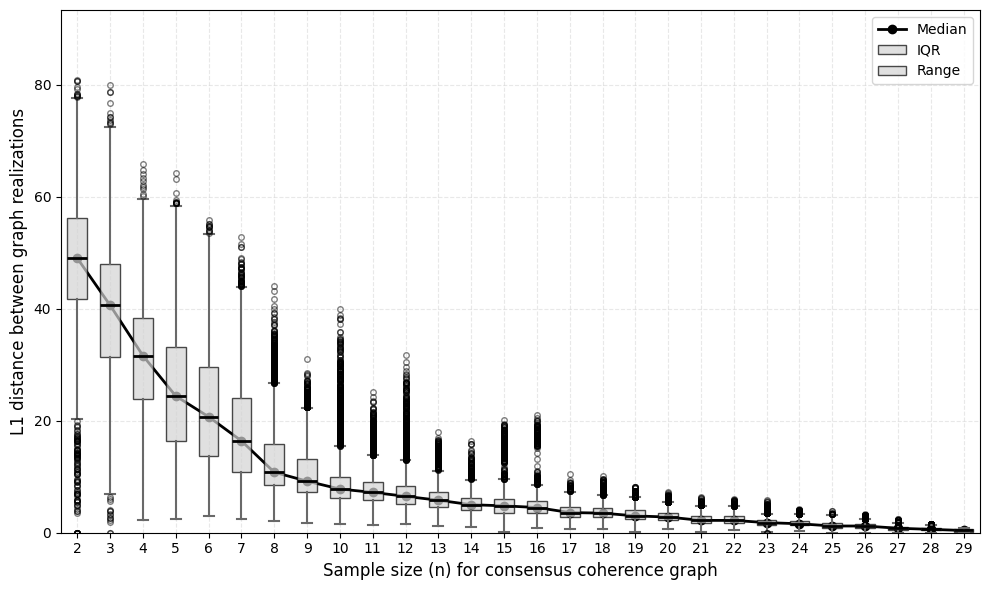}
        \caption{$L^1$/graph edit distance between medians of $n$ of $N = 30$ coherence graphs used to construct the consensus of $N$ coherence graphs in Figure \ref{fig:melianDialogue}. This sort of convergence is typical but not ubiquitous.}
    \label{fig:l1_distances}
\end{figure}

Figure \ref{fig:l1_distances} shows how taking a median of multiple coherence graphs can yield a stable, broadly reproducible result.

\section{\label{sec:prompt_example} Prompt for global consistency inference under uncertainty}

We show an example of the prompt used as a template for all inference attempts in Table \ref{tab:prompt}. 

\begin{table}[ht]
\centering
\begin{tabular}{|p{0.8\textwidth}|}
\hline
For the input set of propositions, identify which propositions are logically consistent
(i.e., can coexist without contradiction). Construct a networkx graph where inconsistent edges
are weight 1 and consistent edges are weight 0. If two vertices do not
involve the same variables, do not create an edge between them.\\
\\
Return solely the edge list with proposition names for vertices. i.e., return responses
in this format:
\\
{[}('b', 'c', 0), \\
('b', 'e', 0), \\
('c', 'd', 0), \\
('c', 'e', 0){]} \\
 \\
Variables: \\
{['$q_1$', '$q_2$', '$q_3$', '$q_4$']} \\
 \\
A given variable can have these properties: \\
property P: P XOR !P \\
property Q: Q XOR !Q \\
property R: R XOR !R \\
property S: S XOR !S \\
 \\
A given property is assigned in a fuzzy manner: \\
property 1: !P := $<$ 0.5P. P := $\geq$ 0.5P \\
property 2: !Q := $<$ 0.5Q. Q := $\geq$ 0.5Q \\
property 3: !R := $<$ 0.5R. R := $\geq$ 0.5R \\
property 4: !S := $<$ 0.5S. S := $\geq$ 0.5S \\

 \\
Input: \\
{[}'Proposition(a): "$q_1$ is !P"', \\
'Proposition(b): "$q_2$ is !P"', \\
'Proposition(c): "$q_1$ is P AND $q_2$ is P AND $q_3$ is P"', \\
'Proposition(d): "$q_4$ is !P"', \\
'Proposition(e): "$q_3$ is !P AND $q_4$ is P"'{]} \\
 \\
\hline
\end{tabular}
\label{tab:prompt}
\caption{Prompt for global consistency inference}
\end{table}

\section{\label{sec:prompt_practical} Prompt for practical examples}

The prompt we used for practical examples such as shown in Figures \ref{fig:melianDialogue} and \ref{fig:minuteMystery} is in Table \ref{tab:prompt_practical}. It evolved from prompts used since spring 2023 to determine models' ability to evaluate the pairwise consistency of propositions \cite{huntsman2024prospects}.

\begin{table}[ht]
\centering
\begin{tabular}{|p{0.8\textwidth}|}
\hline
Imagine that you are a perfectly objective arbitrator with impeccable judgment and integrity. In response to a prompt of the form 'buildCoherence: ' below followed by a list of labeled propositions, please do the following: First, determine which pairs of propositions are substantively related. Second, for each related pair of propositions, determine their logical relationship, assuming that at least one is true, whether or not either actually is. I want you to ignore the truth, falsity or basis in fact of either claim. Third, based on your determination just above, numerically rate the relative consistency of the two propositions. Do not pay attention to or comment on the truth or basis in fact of either proposition independent of the other. Your rating of relative consistency should be on a scale from 0 to 10, with a value of 0 for a pair of propositions that are not at all consistent and a value of 10 for a pair of propositions that are totally consistent. I cannot emphasize enough that for your rating, I want you to ignore the truth or basis in fact of either proposition, since anything that is not consistent with reality cannot be true. If you determine that propositions are unrelated despite previously determining otherwise, omit that pair. To be clear, a pair of false but consistent claims should also be rated a 10. Meanwhile, a pair of propositions of which one is true and the other is false, should be rated a 0. Finally, construct a NetworkX graph where propositions are vertices and edges correspond to substantively related pairs of propositions, with weights given by the consistency ratings just above. Only return the edge list with proposition labels for vertices. i.e., return responses in this format (here 'p2', 'p3', 'p4', and 'p5' are labels): [('p2', 'p3', 0), ('p2', 'p5', 10), ('p3', 'p4', 9), ('p3', 'p5', 2)]. Order vertices (in edges) and edges (in the graph) lexicographically.\\ \\ buildCoherence: \\
\hline
\end{tabular}
\label{tab:prompt_practical}
\caption{Prompt for practical examples}
\end{table}

\section{\label{sec:l1_norm_graph_anova} Fidelity of the reconstruction of the graph measured by the $L^1$ norm normalized by graph size}

The reconstruction of a coherence graph $G_\sigma$ is automatically of the form $G'_{\sigma'}$ where $V(G_\sigma) = V(G'_{\sigma'})$. Therefore, it is easy and appropriate to use the metric $d(G_\sigma,G'_{\sigma'}) := \|A(G_\sigma)-A(G'_{\sigma'})\|_1$, where adjacency matrices are indicated, and we permit $A(\cdot) \in [-1,1]$ here for the sake of generality. This coincides with a graph edit distance in which edge insertion and deletion costs are the absolute value of the edge weight, and in which edge substitution costs are the absolute value of the difference of edge weights. Normalizing this by $|V|\cdot(|V|-1)/2$ gives a size-independent gauge of performance.

\begin{figure}[htbp]
    \centering
    \includegraphics[width=1\linewidth]{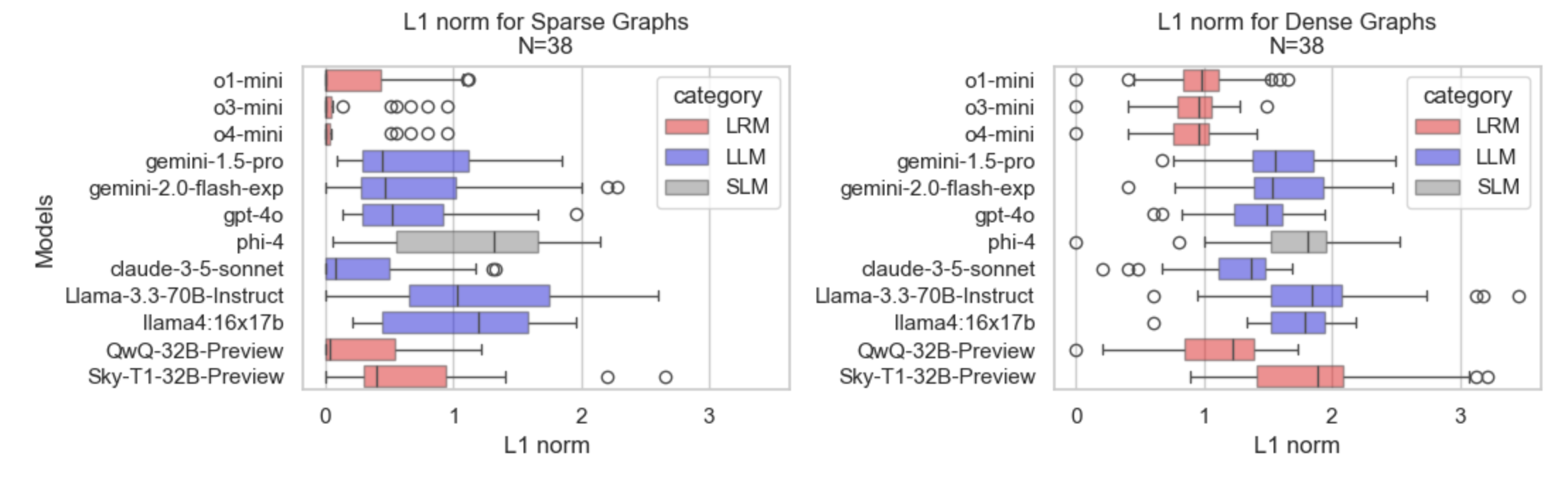}
        \caption{\texttt{o1/3/4-mini}, \texttt{claude-3.5-sonnet}, and \texttt{QwQ-32B} have lower (higher fidelity) $L_1$ scores than other models.}
    \label{fig:cohere_graphs_l1}
\end{figure}

Figure \ref{fig:cohere_graphs_l1} shows that \texttt{o1/3/4-mini}, \texttt{claude-3.5-sonnet} and \texttt{QwQ-32B-Preview} out perform all other models on \texttt{sparse} problems using this measure of adjacency matrix similarity to characterize reconstruction error.

\end{document}